\documentclass[10pt,journal,compsoc]{IEEEtran}
\ifCLASSOPTIONcompsoc
  \usepackage[nocompress]{cite}
\else
  \usepackage{cite}
\fi
\ifCLASSINFOpdf
\else
\fi
\usepackage{graphicx}
\usepackage{amsmath}
\usepackage{amssymb}
\usepackage{booktabs}
\usepackage{multirow}
\usepackage{mathrsfs}
\usepackage{amsfonts}
\usepackage{mathabx}
\usepackage{algorithm}  
\usepackage{algorithmic}

\usepackage{comment}
\usepackage{xcolor}
\newcommand{\myviolet}[1]{{\color{black}{#1}}}

\begin{document}
%
\title{Video Imprint}
%
%
%
%

\author{Zhanning Gao,
        Le Wang,~\IEEEmembership{Member,~IEEE},
        Nebojsa Jojic, 
        Zhenxing Niu,~\IEEEmembership{Member,~IEEE},
        Nanning Zheng,~\IEEEmembership{Fellow,~IEEE},
        Gang Hua,~\IEEEmembership{Senior Member,~IEEE}%
\IEEEcompsocitemizethanks{
{\IEEEcompsocthanksitem Z. Gao, L. Wang and N. Zheng are with the Institute of Artificial Intelligence and Robotics, Xi`an Jiaotong University, Xi`an, Shaanxi 710049, China.
E-mail: zhanninggao@gmail.com, \{lewang, nnzheng\}@mail.xjtu.edu.cn.
\IEEEcompsocthanksitem N. Jojic and G. Hua are with the Microsoft Research, Redmond, Washington, WA 98052.
E-mail: jojic@microsoft.com, ganghua@gmail.com.
\IEEEcompsocthanksitem Z. Niu is with Alibaba Group, Hangzhou, Zhejiang 311121.\protect\\
E-mail: zhenxing.nzx@alibaba-inc.com.}
}
\thanks{Manuscript received 16 Dec. 2017; revised 10 July 2018; accepted 12 Aug. 2018. Date of publication  XX XXX 2018; date of current version  XX XXX 2018.}
\thanks{(Corresponding author: Le Wang.)}
\thanks{Recommended for acceptance by XXX.}
}
%
%

\markboth{IEEE Transactions on Pattern Analysis and Machine Intelligence,~Vol.~XX, No.~XX, XXXX~XXXX}{Shell \MakeLowercase{\textit{et al.}}: Video Imprint}
%



\IEEEtitleabstractindextext{%
\begin{abstract}

\myviolet{A new unified video analytics framework (ER3) is proposed for complex event retrieval, recognition and recounting, based on the proposed video imprint representation, which exploits temporal correlations among image features across video frames. With the video imprint representation, it is convenient to reverse map back to both temporal and spatial locations in video frames, allowing for both key frame identification and key areas localization within each frame. In the proposed framework, a dedicated feature alignment module is incorporated for redundancy removal across frames to produce the tensor representation, {\em i.e.,} the video imprint. Subsequently, the video imprint is individually fed into both a reasoning network and a feature aggregation module, for event recognition/recounting and event retrieval tasks, respectively. Thanks to its attention mechanism inspired by the memory networks used in language modeling, the proposed reasoning network is capable of simultaneous event category recognition and localization of the key pieces of evidence for event recounting. In addition, the latent structure in our reasoning network highlights the areas of the video imprint, which can be directly used for event recounting. With the event retrieval task, the compact video representation aggregated from the video imprint contributes to better retrieval results than existing state-of-the-art methods.}
\end{abstract}

}

\maketitle

\IEEEdisplaynontitleabstractindextext

%
\IEEEpeerreviewmaketitle

\IEEEraisesectionheading{\section{Introduction}\label{sec:introduction}}

%
%
%
%

\IEEEPARstart{A}{nalysis} \myviolet{of event videos is generally considered more challenging than the related video-based action recognition task \cite{caba2015activitynet,simonyan2014two}, thanks to the richer contents of such event videos. Typical event videos are much longer (several minutes or even hours) than trimmed action recognition videos, and multiple human actions and a variety of different objects often appear across various scenes.} For example, a ``birthday party'' event may take place at home or in a restaurant, with multiple objects coming into focus, {\it e.g.}, a birthday cake, and may include a variety of activities that span multiple frames, {\it e.g.}, singing the birthday song, or blowing out candles. 


In the last decade, analysis of complex events in videos has attracted significant attention in the computer vision community \cite{bilen2017action,chang2017semantic,gan2015devnet,gan2016you,jiang2013high,ma2013complex,revaud2013event,sun2013active,tran2014video}. \myviolet{Previous research could be categorized as the unsupervised and the supervised methods. Unsupervised {methods} were typically used for {\it event retrieval} \cite{douze2013stable,revaud2013event} {where the goal is the retrieval of all related videos semantically relevant to the query video sample}. On the other hand, supervised learning has been {widely} used in {\it event recognition} \cite{bhattacharya2014recognition,cao2012scene} and detection tasks \cite{ma2013complex,xu2015discriminative} in similar ways {to its applications in} action recognition \cite{caba2015activitynet,simonyan2014two,zhang2017VA} and {generic video classification tasks} \cite{jiang2015exploiting,wu2015modeling,yue2015beyond}. In the latter case, {a classifier is trained on the annotated training set to recognize the event categories of test videos}, {\it e.g.}, the multimedia event detection task of the TRECVID \cite{over2014trecvid}. In practical applications, it is often desirable to provide explainable results by qualifying the category prediction with the localization of the key pieces of evidence that lead to the recognition decision, which is sometimes referred to as the {\it event recounting}.}

\begin{figure}[t]
\begin{center}
   \includegraphics[width=1\linewidth]{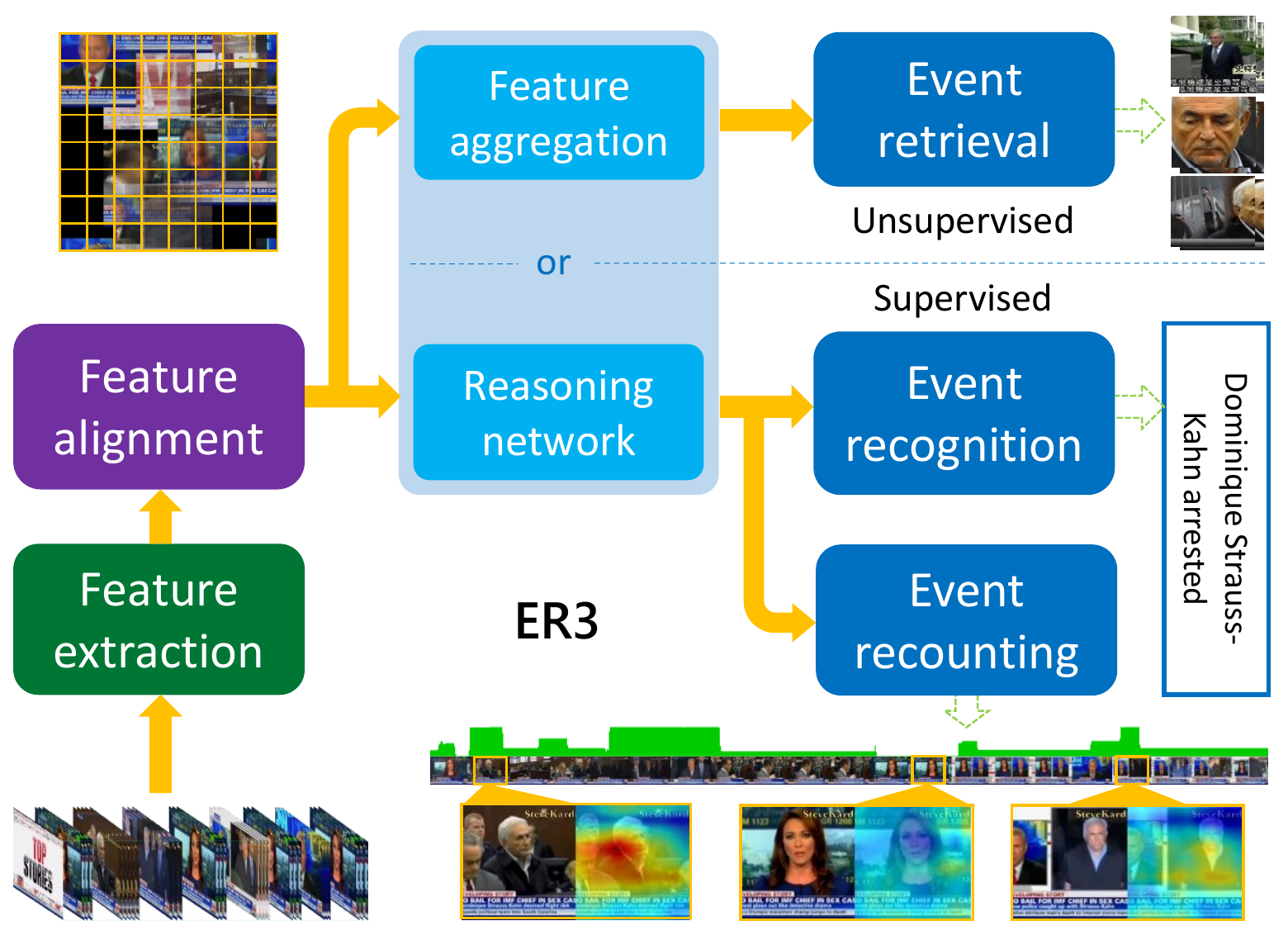}
\end{center}
   \caption{Illustration of the ER3 framework for event retrieval, recognition and recounting. The compact video representation from feature aggregation can be used for large-scale event retrieval. With supervised training, ER3 can also recognize the event category of the input video. Event recounting falls directly out of the latent structure of the model in form of statistics displayed as heat maps for each frame indicating key areas related to the event.}
\label{framework}
\end{figure}

\myviolet{A major challenge in {\it event retrieval} and {\it event recognition} is the construction of appropriate video representations, which should ideally be both discriminative for efficient disambiguation and compact for computational efficiency. Conventionally, a video representation is a fixed-length per-video global feature vector extracted from many frame-level appearance features  \cite{douze2013stable,gan2015devnet,revaud2013event,wu2015modeling,xu2015discriminative}. In a typical {\it event recognition} task, this video representation is fed into a linear classifier \cite{xu2015discriminative} or a neural network \cite{jiang2015exploiting,wu2015modeling} for classification. However, this procedure is generally incompatible with event recounting, as tracing the decision back to individual frame locations is impossible due to the irreversible video representation (global feature) extraction. Therefore, most existing methods perform event recounting as an extra post-processing step \cite{gan2015devnet,lai2014recognizing,tsai2014highly}.}

\myviolet{To address these challenges, the ER3 framework is proposed to simultaneously achieve {\it event retrieval, recognition} and {\it recounting}. Figure \ref{framework} illustrates the components and the inputs/outputs of such ER3 system. } In ER3, (\romannumeral1) we introduce a feature alignment step which can significantly suppress the redundant information and generate a more comprehensive and compact video representation called {\it video imprint}. In addition, the video imprint also preserves the local spatial layout among video frames. (\romannumeral2) Based on the video imprint, in unsupervised setting, we propose an efficient aggregation method for large-scale event retrieval. In supervised setting, we further employ a reasoning network, a modified version of the neural memory networks \cite{sukhbaatar2015end}, which can simultaneously recognize the event category and locate the key pieces of evidence for the event category. In fact, the recounting is so naturally integrated in the framework that the experiments show that the recounting step can assist the recognition task and improve the recognition accuracy. (\romannumeral3) \myviolet{In the recounting task, both temporal key frame identification (attribution of the key frames with respect to the event category, as in \cite{lai2014recognizing,tsai2014highly}) and spatial key areas localization (attribution of the key areas within each frame) are implemented, thanks to the video imprint preserving local semantic and spatial layouts. }


%
\myviolet{This manuscript is an extension of our conference paper \cite{gao2017er3} with modifications as follows. In the feature alignment step, an alternative generative model ({\it i.e.}, epitome \cite{jojic2003epitomic}) is included besides the the tessellated counting grid (TCG) model \cite{perina2011image,perina2015capturing}. Both generative models share the core idea of building condensed representation by exploiting the spatial interdependence among the input features. However, unlike the TCG model which is limited to counts/histogram-style input features, the epitome model is represented as structured Gaussian mixtures which can accommodate general vector or tensor input features.  In addition, to accelerate the feature alignment step with the epitome model, we propose an accelerated two-step scheme to update the epitome. More details are provided in Section \ref{alignmentStep}. We also provide a comprehensive comparison between the two generative models with various datasets and tasks. The experimental results show that the alternative epitome model achieves higher computational efficiency with comparable results with the TCG model. }

The paper is organized as follows. Section \ref{RelatedWork} discusses related work about event videos analysis. Then, we present the technical details of the ER3 in Section \ref{ER3}. Experimental results are provided in Section \ref{Experiments}. Finally, we conclude the paper in Section \ref{Conclusion}.

\begin{figure*}[t]
\begin{center}
   \includegraphics[width=1\linewidth]{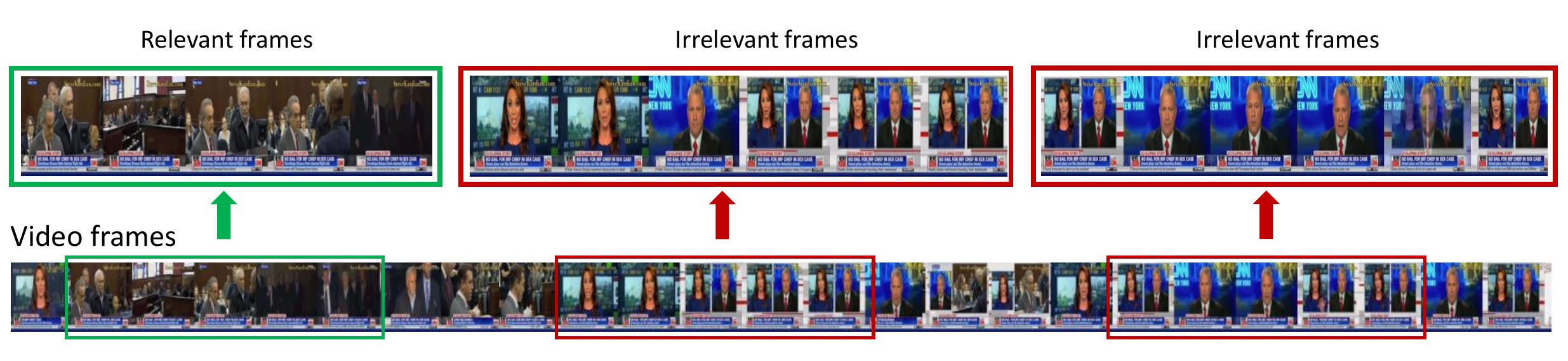}
\end{center}
   \caption{Illustration of the frames related to ``Dominique Strauss-Kahn arrested''. The frames in green box denote the positive frames related to the event. Red boxes show irrelevant frames.}
\label{figure_overcount}
\end{figure*}

\section{Related work}\label{RelatedWork}


\myviolet{With a typical unsupervised event retrieval task, the goal is the retrieval of all related videos semantically relevant to the query video sample. A major challenge is the construction of both compact and discriminative video representations. Conventional methods \cite{douze2013stable,poullot2015temporal,revaud2013event} rely on frame-level local features ({\em e.g.}, SIFT~\cite{lowe2004distinctive}) and aggregation strategies ({\em e.g.}, Fisher Vector~\cite{sanchez2013image,perronnin2015fisher}, VLAD~\cite{douze2013stable,jegou2010aggregating,jegou2012aggregating} or explicit feature maps \cite{poullot2015temporal}) for frame-level feature description. } \myviolet{Recent works \cite{jiang2015exploiting,xu2015discriminative} predominantly employ deep Convolutional Neural Network (CNNs) \cite{he2016resnet,simonyan2014very} to extract a feature descriptor from each video frame. Latest work \cite{baraldi2018lamv} revisits temporal match kernels \cite{poullot2015temporal} and presents a learnable temporal layer which can further enhance the CNNs based feature descriptor. Subsequently, a video-level representation is typically obtained by directly averaging all frame-level descriptors in the video. Such sum-aggregation strategy disregards the strong temporal correlations among consecutive frames, which may over-emphasize certain long or recurrent shots in the video.} We discuss this problem in Section \ref{ER3} and show that the redundant information among frames can be effectively suppressed by the feature alignment step. 

\myviolet{Event recognition and detection have attracted wide attention in the last decade. In general, event video recognition system usually consists of three stages, {\it i.e.}, feature extraction, feature aggregation/pooling and training/recognition. As in event retrieval, the first two stages aim at building discriminative video representations. Previous methods focused mostly on designing better video features or representations for the classifier, such as hand-crafted visual features \cite{dalal2005histograms,lowe2004distinctive}, motion features \cite{wang2011action,wang2013action}, audio features \cite{baillie2003audio}, and mid-level concept/attribute features \cite{chang2015searching,torresani2010efficient}. Recently, the advancements of CNN \cite{krizhevsky2012imagenet,simonyan2014very} lead to promising results in event recognition task \cite{jiang2015exploiting,xu2015discriminative,BMVC2015_60}. The video representations are usually constructed by direct aggregation of the frame-level CNN features. Due to limited amount of training data, video representations are typically fed to classifiers such as the Support Vector Machine (SVM)\cite{chang2011libsvm} or multi-feature fusion framework \cite{jiang2015exploiting,BMVC2015_60,wu2015modeling,zhang2015multi}.}

Event recounting refers to the attribution of key pieces of evidence supporting the recognition decision. As with most video analytics datasets, only video-level annotations are provided, making such attribution a challenging task. Event recounting is usually implemented as a post-processing step after the recognition \cite{lai2014recognizing,tsai2014highly}. Chang {\it et al.} \cite{chang2015searching} proposed a joint optimization framework with mid-level semantic concept representations for event recognition and recounting. Sun {\it et al.} \cite{sun2014discover} introduced an evidence localization model learned via a max-margin framework, and Lai {\it et al.} \cite{lai2014video} applied Multiple-Instance Learning (MIL) which infers temporal instance labels and the video-level labels. In these works, event videos are treated as sets of shots or instances, these recounting procedures perform only temporal localization and usually at a coarse scale. Recently, Gan {\it et al.} \cite{gan2015devnet} proposed a deep neural network for event recognition. Specifically, event recounting (both temporal and spatial) is also achieved via passing the classification scores backward. However, this is still an explanatory post-processing step which is never designed to assist the event recognition task.

In contrast with these methods, at the core of our integrated event recognition and recounting framework is a latent structure that contains reverse attribution pointers back to the video frames. In addition, the proposed framework is trained in an unsupervised manner by simultaneously aligning areas across frames and estimating a probability distribution over frame features in the corresponding areas. The obtained representation consists of a grid of distributions with the corresponding mappings from the feature to the frames, similar to the way video frames are mapped to panoramas from pixel space (see the toy examples in Figure \ref{framework} and Figure \ref{figure_alignment}). The obtained grid and such reverse mapping pointers (back to the spatial locations of the specific video frames) form the proposed \emph{video imprint}. With this video imprint representation, it is possible to design an aggregation to emphasize mere presence instead of frequency of repetition.

\myviolet{Our experiments show that the proposed video imprint aggregation yields better performances in both the supervised and unsupervised tasks than existing algorithms. The video imprint also allows for the reasoning over the spatial layouts of features across frames. Inspired by the attention mechanism in memory networks \cite{sukhbaatar2015end} reasoning over sentences in priming text and video face recognition \cite{yang2017neural}, the proposed reasoning network analyzes evidences at different spatial locations of the compact video imprint while carrying out the recognition task, which also highlights the key areas of the imprint. These key areas of the imprint could readily be mapped back to specific video frames and corresponding spatial locations. In this way, event recounting is implemented as an integral part of recognition instead of a post-processing step.}

\section{The details of ER3}\label{ER3}

In this section, we present all modules and the operating mechanism of the proposed ER3 framework, as illustrated in Figure \ref{framework}.

\subsection{Feature extraction}

\myviolet{Recently, image descriptors based on the activations of convolutional layers \cite{gong2014multi,sharif2014cnn,BMVC2015_60} have outperformed previous methods \cite{jiang2015exploiting,BMVC2015_60} based on features extracted from fully connected layers. Inspired by the spatial information preserving characteristics of convolutional layers outputs, we also choose the activations of the last convolutional layer as the frame-level feature.}

\subsection{Feature alignment}\label{alignmentStep}

\myviolet{Existing event recognition algorithms rely on a compact and discriminative video representation, which is typically direct average of the frame-level descriptors \cite{douze2013stable,jiang2015exploiting,xu2015discriminative,BMVC2015_60}. Such sum-aggregation strategy disregards the strong temporal correlations among consecutive frames, which may over-emphasize certain long or recurrent shots in the video. Therefore, irrelevant and repetitive shots in the video might dominate the obtained video representation. For instance, in the event ``Dominique Strauss-Kahn arrested" as shown in Figure~\ref{figure_overcount}, many video frames showing the news anchor are irrelevant, but they are visually similar, therefore the simple averaging aggregation strategy for video representation may over-emphasize such irrelevant frames. To mitigate this problem, we propose the feature alignment procedure to regularize the influence of frame features.}


\myviolet{The idea of feature alignment is inspired by panoramic stitching \cite{szeliski2006image,szeliski1997creating}, which can remove the redundant or overlapping parts between multiple images. The redundancy across video frames could likewise be significantly reduced to improve the robustness of the obtained video representation against long and/or repetitive irrelevant video frames.}

\myviolet{Due to the dynamic and complex nature of event videos, it is impractical to directly stitch video frames by pixel-level alignment. Instead, image features are extracted first using the activations of the last convolutional layer with each frame as input. Afterwards, the tessellated counting grid (TCG) \cite{perina2011image,perina2015capturing} or epitome model \cite{jojic2003epitomic} is employed to generate a tensor of frame-level feature distributions, which implies a panorama-style reversible mapping, accommodating the geometric variations in objects and scenes in event videos. This model exploits the spatial interdependence of the frame-level features in relevant frames, which makes possible the subsequent clustering of visually similar shots. }

\myviolet{The inquisition for leveraging the epitome model \cite{jojic2003epitomic} is to accelerate the feature alignment step. Both the TCG and the epitome can capture the spatial interdependence among input features. However, the input features of the epitome model can be more flexible, considering their Gaussian location distribution instead of the discrete categorical distribution in the TCG. In addition, the incorporation of the epitome model also allows us to propose a highly efficient two-step implementation, which leads to approximately an order of magnitude faster feature alignment procedure than the counterpart in the TCG. In the remainder of this section, mathematical notations are first summarized in Table \ref{table_notations}, followed by introductions of both generative models \footnote{Comprehensive introductions to both the TCG and the epitome models can be find in \cite{perina2011image,perina2015capturing,jojic2003epitomic}.}.  }

\setlength{\tabcolsep}{6pt}
\begin{table}[t]
\begin{center}
\caption{\myviolet{Principal Notations.}}

\label{table_notations}
\begin{tabular}{lp{7cm}}
\toprule
${\boldsymbol \pi}_{\bf i}$   & Parameters of the counting grid model. $l_1-$normalized counts feature at location ${\bf i}$ on the counting grid ${\bf{E}}=[1 \dots E_x] \times [1 \dots E_y]$\\
$\pi_{{\bf i},z}$             & $z$-th dimension of ${\boldsymbol \pi}_{\bf i}$ and $\sum_{z \in {\bf Z}}{\pi_{{\bf i},z}} = 1$, where $z \in {\bf Z} = [1, \dots, Z]$ \\
$({\boldsymbol \mu}_{\bf i},{\boldsymbol \sigma}^2_{\bf i})$   & Parameters of the epitome model. The mean ${\boldsymbol \mu}_{\bf i}$ and variance ${\boldsymbol \sigma}^2_{\bf i}$ of the Gaussian distribution aligned at the location $\bf i$ of the grid ${\bf E}$\\
$\{{\bf c}^s\}_{s \in {\bf S}}$ & Counts features plugged in a tessellation ${\bf S}$, where ${\bf s \in S} = [1 \dots S_x] \times [1 \dots S_y]$ \\
${\bf F}$                     & Tensor features, {\it e.g.}, the activations from convolutional layer of the CNN model\\
${\boldsymbol f}_{\bf j}$     & The feature vector  extracted  from ${\bf F}$ along  the  spatial  dimensions, {\it i.e.}, ${\bf j \in W} = [1,\dots,W_x]\times[1,\dots,W_y]$\\
${\bf W_k}$                   & The window at the location ${\bf k}$ of the grid ${\bf E}$ which assumed to generate counts features $\{{\bf c}^s\}_{s \in {\bf S}}$ or tensor features ${\bf F}$\\
${\bf W^s_k}$                 & The sub-window in the ${\bf W_k}$ generating each ${\bf c^s}$\\
$l$                           & The latent variable that represents the mapping location in the grid ${\bf E}$\\

\bottomrule
\end{tabular}

\end{center}
\end{table}

\begin{figure}[t]
\begin{center}
   \includegraphics[width=0.95\linewidth]{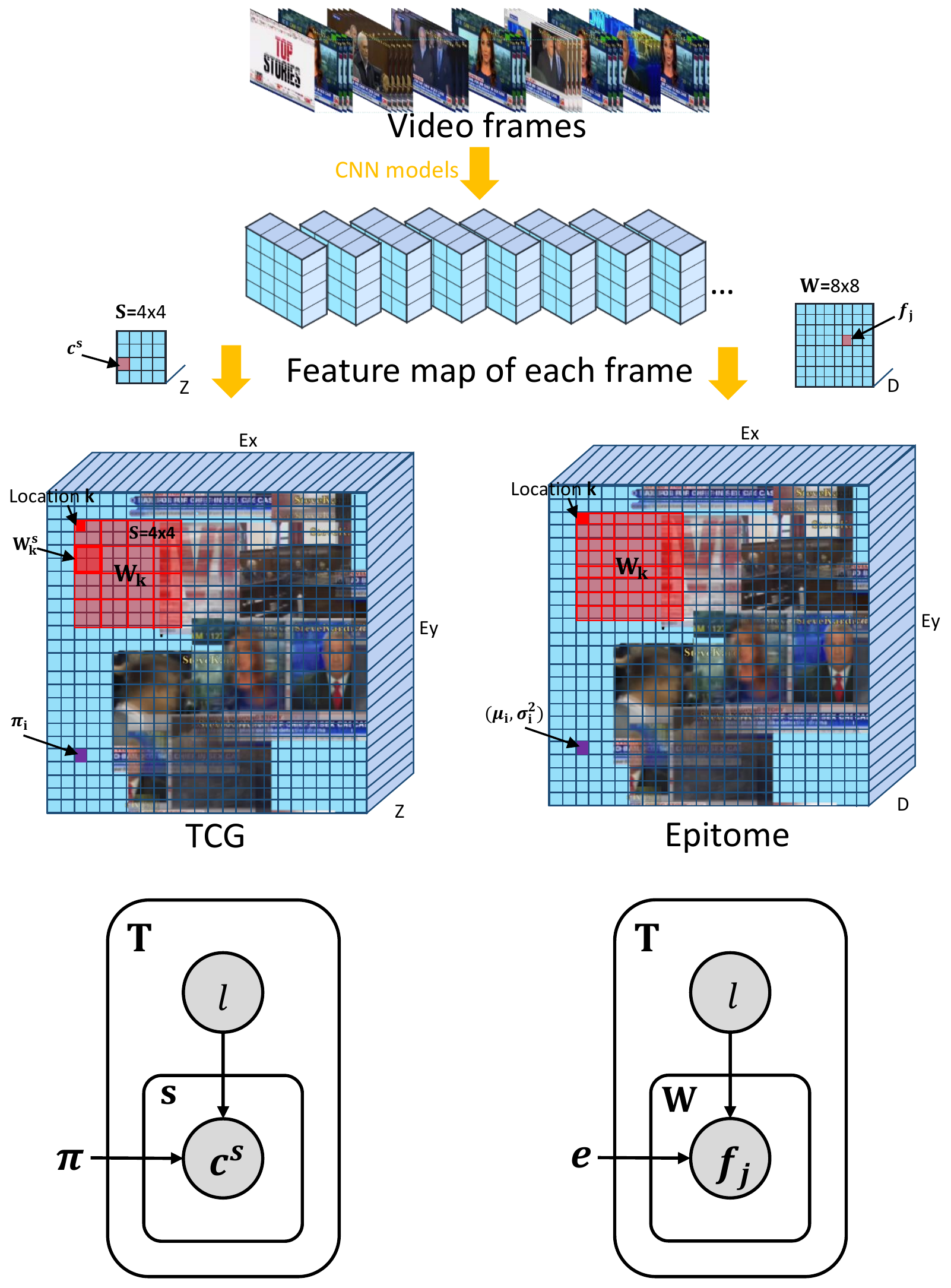}
\end{center}
   \caption{\myviolet{Illustration of tessellated counting grid (TCG) and epitome models, and their Bayesian networks. The left tensor block represents the TCG with ${\bf E}=24\times 24, {\bf W}=8\times 8, {\bf S} = 4\times 4$. The right tensor block represents the epitome with ${\bf E}=24\times 24, {\bf W}=8\times 8$. For TCG, the input CNN feature maps are down-sampled to ${\bf S} = 4\times 4$. In fact, the epitome can be regarded as a special case of TCG when ${\bf W=S}$. For both TCG and epitome, similar frames are usually represented in the same or nearby windows, {\it e.g.}, the anchor who we frequently see in the video.}}
\label{figure_alignment}
\end{figure}

\subsubsection{Tessellated counting grid}

{Tessellated counting grid} (TCG) \cite{perina2011image,perina2015capturing} is designed to capture the spatial interdependence among image features. \myviolet{Given a set of images or a video sequence, it assumes that each image/frame is represented by a set of $l_1-$normalized non-negative feature vectors ({\it e.g.}, bags of visual words vectors) $\{{\bf c}^s\}_{s \in {\bf S}}$ plugged in a tessellation ${\bf S} = [1, \dots, S_x] \times [1, \dots, S_y]$ }{\footnote {With $l_1-$normalization and appropriate down-sampling, the feature maps (after ReLU) from the convolutional layer of CNN model naturally satisfy this assumption.}}. 

\myviolet{Formally, the counting grid ${\boldsymbol \pi}_{{\bf{i}}}$ is a set of normalized counts of features indexed by $z \in {\bf Z} = [1 \dots Z]$ (dimension of image feature) on the 2D discrete grid ${\bf{i}}=(i_x,i_y) \in {\bf{E}}=[1, \dots, E_x] \times [1, \dots, E_y]$, where $\bf{i}$ denotes the location on the grid and $\sum_{z \in {\bf Z}}{\pi_{{\bf i},z}} = 1$  \cite{jojic2011multidimensional}.} 

As a generative model, the image features $\{{\bf c}^s\}_{s \in {\bf S}}$ are assumed to follow a distribution found in a window into the counting grid. \myviolet{The probability of generating the image features $\{{\bf c}^s\}_{s \in {\bf S}}$ from the window ${\bf W}_{\bf{k}} = [k_x, \dots, k_x + W_x - 1]\times[k_y, \dots, k_y + W_y - 1]$ placed at the location ${\bf{k}} = (k_x, k_y) \in {\bf E}$ of the grid is}
\begin{equation}
p(\{{\bf c}^s\}_{s \in {\bf S}}|l={\bf{k}}) = \gamma \prod_{z \in {\bf Z}}{\prod_{s \in {\bf S}}\left(\sum_{{\bf{i}}\in{\bf{W}}_{\bf k}^s}{\pi_{{\bf{i}},z}}\right)^{c_z^s}},
\end{equation}
where $\gamma$ is the normalization constant. \myviolet{$l$ denotes the latent variable, while ${\bf i}$ and ${\bf k}$ represent generic positions in the grid ${\bf E}$.} Then, for a given counting grid ${\boldsymbol \pi}$, the joint distribution over the set of image features $\{{\bf c}^{s,t}\}_{s \in {\bf S},t \in {\bf T}}$, indexed by $t \in {\bf T} = [1 \dots T]$, and their corresponding latent window locations $\{l^t\}$ in the grid can be derived as
\begin{equation}
P(\{{\bf{c}}^{s,t}\}_{s \in {\bf S},t \in {\bf T}},\{l^t\}_{t \in {\bf T}}) \propto \prod_{t \in {\bf T}}\sum_{{\bf{k} \in {\bf E}}}\prod_{z \in {\bf Z}}\prod_{s \in {\bf S}}{\left(\sum_{{\bf{i}}\in{\bf{W}}_{\bf k}^s}{\pi_{{\bf{i}},z}}\right)^{c_z^{s,t}}}.
\end{equation}

The counting grid $\boldsymbol{\pi}$ can be estimated by maximizing the log likelihood of the joint distribution with an EM algorithm,
\begin{equation}
\label{eq3}
\begin{aligned}
{\rm{E}} \ {\rm{step:}}\ \ & q(l^t={\bf{k}}) \propto \exp\left(\sum_{s \in {\bf S}}\sum_{z \in {\bf Z}}{c_z^{s,t}}\log \sum_{{\bf{i}}\in{\bf{W}}_{\bf k}^s}{\pi_{{\bf{i}},z}}\right), \\
{\rm{M}} \ {\rm{step:}}\ \ & {\pi_{{\bf{i}},z}} \propto {\pi_{{\bf{i}},z}^{old}}\sum_{t \in {\bf T}}{\sum_{s \in {\bf S}}{c_z^{s,t}}\sum_{{\bf{k}}|{\bf{i}}\in{{\bf{W}}_{\bf{k}}^s}}{\frac{q(l^t={\bf{k}})}{\sum_{{\bf{i}}\in{\bf{W}}_{\bf{k}}^s}{\pi_{{\bf{i}},z}^{old}}}}},
\end{aligned}
\end{equation}
where $q(l^t={\bf{k}})$ denotes the posterior probability $p(l^t={\bf{k}}|\{{\bf c}^{s,t}\}_{s \in {\bf S}})$ and ${\pi_{{\bf{i}},z}^{old}}$ denotes the counting grid at the previous iteration. 


\myviolet{The iterative process of TCG jointly estimates the counting grid ({\em i.e.},  \emph{video imprint}) $\boldsymbol{\pi}$ and aligns all video frame features to it with such correspondences captured in $q$.}

\subsubsection{Epitome}

The original epitome model \cite{jojic2003epitomic} takes raw pixels as input and aims to mine the essence of the textural and shape properties of the image. \myviolet{Formally, the epitome ${\bf e}$ is a set of dependent Gaussian distributions $\{\mathcal{N}({\boldsymbol f}_{\bf j};{\boldsymbol \mu}_{\bf i},{\boldsymbol \sigma}^2_{\bf i})\}$ aligned on a grid ${\bf{i}}=(i_x,i_y) \in {\bf{E}}=[1, \dots, E_x] \times [1, \dots, E_y]$ just like TCG. In the original epitome formulation, ${\boldsymbol f}_{\bf j}$ denotes the intensity or the color of the pixel on the image patch ${\bf F}$ indexed by ${\bf{j}}=(j_x,j_y) \in {\bf{W}}=[1, \dots, W_x] \times [1, \dots, W_y]$. Here we extend ${\bf F}$ to a general tensor (feature map), specifically, the activations from the last convolutional layer of a CNN model. In other words, ${\bf F}$ is the CNN feature map and ${\boldsymbol f}_{\bf j}$ is the feature vector extracted from ${\bf F}$ along the spatial dimensions, {\it i.e.}, ${\bf j} \in {\bf{W}}$.}

Given the epitome ${\bf e}$, the probability of generating the feature map ${\bf F}$ from the window ${\bf W_k}$ at location ${\bf k}$ of the epitome ${\bf e}$ is

\begin{equation}
\label{eq_ep_generator}
p({\bf F}|l={\bf{k}}) = \gamma \prod_{{\bf{i}}\in{\bf{W}}_{\bf k}}\mathcal{N}\left({\boldsymbol f}_{\bf i-k};\boldsymbol \mu_{\bf i},\boldsymbol \sigma^2_{\bf i}\right),
\end{equation}
where $\gamma$ is the normalization constant. $\mathcal{N}\left({\boldsymbol f}_{\bf i-k};\boldsymbol \mu_{\bf i},\boldsymbol \sigma^2_{\bf i}\right)$ is a Gaussian distribution over ${\boldsymbol f}_{\bf i-k}$ with mean $\boldsymbol \mu_{\bf i}$ and variance $\boldsymbol \sigma^2_{\bf i}$ and ${\bf i-k}=(i_x-k_x+1,i_y-k_y+1)$. Similar to TCG, the joint distribution over the set of feature maps $\{{\bf F}^{t}\}_{t \in {\bf T}}$, indexed by $t$, and their corresponding latent window locations $\{l^t\}_{t \in {\bf T}}$ on the epitome can be derived as
\begin{equation}
P(\{{\bf{F}}^{t}\},\{l^t\}) \propto \prod_{t \in {\bf T}}\sum_{{\bf{k \in E}}}\prod_{{\bf{i}}\in{\bf{W}}_{\bf k}}\mathcal{N}\left({\boldsymbol f}^t_{\bf i-k};\boldsymbol \mu_{\bf i},\boldsymbol \sigma^2_{\bf i}\right).
\end{equation}

The parameters $({\boldsymbol \mu, \boldsymbol \sigma^2})$ are estimated by marginalizing the joint distribution, {\it i.e.}, optimizing the log likelihood of the data with an iterative EM algorithm,
\begin{equation}
\label{eq_ep_EM}
\begin{aligned}
{\rm{E}} \ {\rm{step:}}\ \ & q(l^t={\bf{k}}) \propto \prod_{{\bf{i}}\in{\bf{W}}_{\bf k}}\mathcal{N}\left({\boldsymbol f}^t_{\bf i-k};\boldsymbol \mu_{\bf i},\boldsymbol \sigma^2_{\bf i}\right), \\
{\rm{M}} \ {\rm{step:}}\ \ & \boldsymbol \mu_{\bf i} = \frac{\sum_{t \in {\bf T}}{\sum_{\bf k \in W_{i-W}}{q(l^t={\bf{k}})}{\boldsymbol f}^t_{\bf i-k}}}{\sum_{t \in {\bf T}}{\sum_{\bf k \in W_{i-W}}{q(l^t={\bf{k}})}}}, \\
                           & \boldsymbol \sigma^2_{\bf i} = \frac{\sum_{t \in {\bf T}}{\sum_{\bf k \in W_{i-W}}{q(l^t={\bf{k}})}({\boldsymbol f}^t_{\bf i-k}-{\boldsymbol \mu}_{\bf i})^2}}{\sum_{t \in {\bf T}}{\sum_{\bf k \in W_{i-W}}{q(l^t={\bf{k}})}}},
\end{aligned}
\end{equation}
\myviolet{where $q(l^t={\bf{k}})$ denotes the posterior probability $p(l^t={\bf{k}}|{\bf F}^t)$ and ${\bf i-W}=(i_x-W_x+1,i_y-W_y+1)$.}

\myviolet{As presented above, both generative models (TCG and epitome) build condensed representations by capturing the spatial interdependence among input features. The epitome model differs from the TCG in the location distributions (Gaussian versus discrete categorical).} Therefore, the input features of the epitome can be more flexible. See Figure \ref{figure_alignment} for the illustration of the TCG and epitome\footnote{\myviolet{For ease of illustration of the video imprint, we accumulate the frames on the location with the maximum posterior probability $q(l^t={\bf{k}})$ and draw the mean image.}}. In addition, according to Equation (\ref{eq_ep_EM}), the updating of the epitome parameters, {\it i.e.}, the M step, only depends on the current $q$ distribution, {\it i.e.}, the current E step, and the input features. This motivates us to propose an efficient two-step scheme to generate the epitome based video imprint, {\it i.e.}, we can first efficiently estimate the final $q$ distribution, and then calculate the epitome based video imprint directly with Equation (\ref{eq_ep_EM}). The following are the details of the efficient two-step implementation.

\begin{figure*}[t]
\begin{center}
   \includegraphics[width=0.99\linewidth]{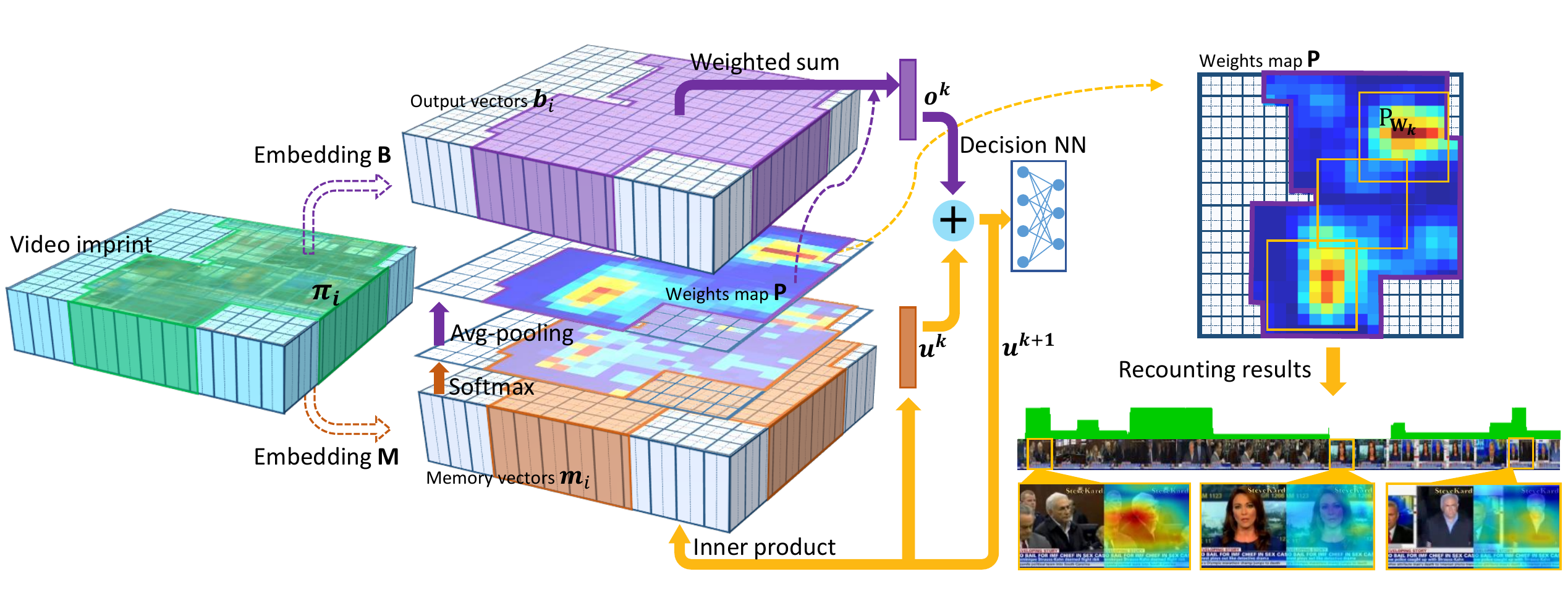}
\end{center}
   \caption{Illustration of the reasoning network for event recognition and event recounting.}
\label{figure_spatialMNN}
\end{figure*}

\subsubsection{An efficient two-step scheme}

Given a fixed epitome size ${\bf E}$ and window size ${\bf W}$, the only feasible way to accelerate the learning procedure of epitome model is to reduce the dimension of the input feature maps $\{{\bf F}^t\}_{t \in {\bf T}}$. {Although the low dimensional input features} produced from $\{{\bf F}^t\}_{t \in {\bf T}}$ by dimension reduction are enough to capture the correspondences among frames within a video, it may also reduce the discriminative capacity of the epitome based video imprint ${\boldsymbol \mu}$ for large-scale video event analysis. Therefore, we propose an efficient two-step scheme which divides the epitome based feature alignment operation into two steps: the correspondence analysis step and the video imprint generation step.

\myviolet{At the first step, the low dimensional feature maps $\{{\bf G}^t\}_{t \in {\bf T}}$ are generated by PCA~\cite{jegou2012negative} projection, ${\bf G} = \phi_{PCA}({\bf F})$, where ${\bf G} \in \mathbb{R}^{W_x \times W_y \times d}, {\bf F} \in \mathbb{R}^{W_x \times W_y \times D}$ and $d \ll D$. Then, the epitome of $\{{\bf G}^t\}_{t \in {\bf T}}$ can be learned with}
\begin{equation}
\label{eq_ep_EM_low}
\begin{aligned}
{\rm{E}} \ {\rm{step:}}\ \ & \hat{q}(l^t={\bf{k}}) \propto \prod_{{\bf{i}}\in{\bf{W}}_{\bf k}}\mathcal{N}\left({\boldsymbol g}^{t}_{\bf i-k};\hat{\boldsymbol \mu}_{\bf i},\hat{\boldsymbol \sigma}^2_{\bf i}\right), \\
{\rm{M}} \ {\rm{step:}}\ \ & \hat{\boldsymbol \mu}_{\bf i} = \frac{\sum_{t \in {\bf T}}{\sum_{\bf k \in W_{i-W}}{\hat{q}(l^t={\bf{k}})}{\boldsymbol g}^t_{\bf i-k}}}{\sum_{t \in {\bf T}}{\sum_{\bf k \in W_{i-W}}{\hat{q}(l^t={\bf{k}})}}}, \\
                           & \hat{\boldsymbol \sigma}^2_{\bf i} = \frac{\sum_{t \in {\bf T}}{\sum_{\bf k \in W_{i-W}}{\hat{q}(l^t={\bf{k}})}({\boldsymbol g}^t_{\bf i-k}-\hat{\boldsymbol \mu}_{\bf i})^2}}{\sum_{t \in {\bf T}}{\sum_{\bf k \in W_{i-W}}{\hat{q}(l^t={\bf{k}})}}}.
\end{aligned}
\end{equation}
As an approximation of the $q$ distribution, we found that $\hat{q}$ can fully capture the correspondence of video frames (see Figure \ref{figure_mAPandTime}). In addition, since $d \ll D$, $\hat{q}$ estimation with Equation (\ref{eq_ep_EM_low}) achieves much higher efficiency.

\begin{algorithm}[t]
\caption{The efficient two-step scheme}  
\label{alg1}  
\begin{algorithmic}[1]
\REQUIRE the feature maps $\{{\bf f}^t\}_{t \in {\bf T}}$, epitome size ${\bf E}$, window size ${\bf W}$
\ENSURE $\hat{q}$, $\boldsymbol \mu$
\FOR{each $t = 1\ ...\ T$}
\STATE ${\bf G}^t = \phi_{PCA}({\bf F}^t)$
\ENDFOR
\REPEAT
\STATE Update $\hat{q}$ with the EM steps of Equation (\ref{eq_ep_EM_low})
\UNTIL{Convergence}
\STATE Compute $\boldsymbol \mu$ with Equation (\ref{eq_ep_EM_final})
\STATE Return $\hat{q}$, $\boldsymbol \mu$
\end{algorithmic}  
\end{algorithm}  

At the second step, according to the M step of the Equation (\ref{eq_ep_EM}), the updated parameters $\boldsymbol\mu$ and $\boldsymbol\sigma^2$ only depend on the set of input feature maps $\{{\bf F}^t\}_{t \in T}$ and the $q$ distribution from the last iteration of E step. Therefore, if we replace $q$ with its approximation $\hat{q}$ from the first step, $\boldsymbol\mu$ and $\boldsymbol\sigma^2$ can be directly computed with
\begin{equation}
\label{eq_ep_EM_final}
\begin{aligned}
&{\boldsymbol \mu}_{\bf i} = \frac{\sum_{t \in {\bf T}}{\sum_{\bf k \in W_{i-W}}{\hat{q}(l^t={\bf{k}})}{\boldsymbol f}^t_{\bf i-k}}}{\sum_{t \in {\bf T}}{\sum_{\bf k \in W_{i-W}}{\hat{q}(l^t={\bf{k}})}}}, \\
&{\boldsymbol \sigma}^2_{\bf i} = \frac{\sum_{t \in {\bf T}}{\sum_{\bf k \in W_{i-W}}{\hat{q}(l^t={\bf{k}})}({\boldsymbol f}^t_{\bf i-k}-{\boldsymbol \mu}_{\bf i})^2}}{\sum_{t \in {\bf T}}{\sum_{\bf k \in W_{i-W}}{\hat{q}(l^t={\bf{k}})}}}.
\end{aligned}
\end{equation}

The computational complexity and performance of the efficient two-step scheme will be discussed in Section~\ref{Experiments}. We shall note that {this fast implementation scheme is inapplicable to TCG.} It is difficult to transfer a high dimensional histogram vector to a low dimensional space. Most importantly, as shown in Equation (\ref{eq3}), the current ${\boldsymbol \pi_i}$ relies on both ${\boldsymbol\pi}_{\bf i}^{old}$ and the $q$ distribution at the previous iteration which makes the efficient two-step scheme invalid for TCG. In practice, $\hat{\boldsymbol\sigma}^2_{\bf i}$ is fixed as 0.1 to avoid overfitting during the EM iterations and only $\mu$ is taken as video imprint which has the same size with $\boldsymbol\pi$. The efficient two-step scheme is summarized in Algorithm \ref{alg1}.

\subsection{Feature aggregation}

In this section, we demonstrate how to aggregate the video imprint into a compact video representation for unsupervised event retrieval. We refer each $\boldsymbol\pi_{\bf i}$ or $\boldsymbol\mu_{\bf i}$ on the video imprint as an imprint descriptor. As shown in Figure \ref{figure_alignment}, some imprint descriptors are meaningless since no frames are aligned to their locations. \myviolet{The first step is to generate an active map for the video imprint to eliminate these meaningless imprint descriptors}. Formally, the binary active map, ${\bf{A}}=\{a_{\bf{i}}|{\bf{i}}\in{\bf{E}}\},a_{\bf{i}}\in\{0,1\}$, is computed as
 \begin{equation}
 \label{eq_active}
 a_{\bf{i}}=\left\{
\begin{aligned}
1 &   & \left\{{\bf i} \in {\bf W_k} |~{\bf k}: \sum_{t=1}^{T}{q(l_t={\bf{k}})} > \tau\right\} \\
0 &   & \rm{otherwise}
\end{aligned},
\right.
\end{equation}
where $\tau$ is the threshold of the active map.

\myviolet{After the active map is obtained, a direct sum-aggregation over the entire activated imprint descriptors is carried out to produce the final video representation.} Formally, the aggregation step can be written as
\begin{equation}
\begin{aligned}
{\rm TCG:}\ \ &\phi_{FA}(\boldsymbol\pi,{\bf{A}})=\sum_{{\bf{i\in{E}}}}{a_{\bf{i}}{\boldsymbol\pi_{\bf{i}}}}, \\ 
{\rm Epitome:}\ \ &\phi_{FA}(\boldsymbol\mu,{\bf{A}})=\sum_{{\bf{i\in{E}}}}{a_{\bf{i}}{\boldsymbol\mu_{\bf{i}}}}.
\end{aligned}
\label{FA}
\end{equation}

The obtained $\phi_{FA}$ is subsequently $l_2$-normalized and the cosine similarity is computed for event retrieval.

\subsection{Reasoning over the imprint}


\myviolet{Once the imprint is obtained for each video, it makes reasoning based on this compact imprint representation possible, where each location corresponds to a recurring scene/object part, with spatial layouts of imprint locations reflecting the scene/object spatial layouts in video frames where they were observed. We treat locations in the imprint in a similar way to the sentences in memory networks \cite{weston2015towards,sukhbaatar2015end}. Our reasoning network determines the event category in stages, which traverses attention from one set of imprint locations to the next. In this process as illustrated in Figure \ref{figure_spatialMNN}, the imprint locations of large importance are highlighted, and we also trace these highlights back to the locations in video frames using the $q/\hat{q}$ distributions as discussed above. }

Our reasoning network differs from the memory networks in two ways. First, since there is no query question for event recognition, we initialize the input vector ${u}^1$ with Equation (\ref{FA}), {\it i.e.}, sum-aggregation of video imprint. Second, because the spatial organization in the imprint is meaningful, \myviolet{we incorporate an average spatial pooling layer after the softmax layer to improve the smoothness of the recounting results}. The model details are as follows. 

{\bf Memory layers in the reasoning network}. As shown in Figure \ref{figure_spatialMNN}, the video imprint (non-activated locations are ignored) is processed via multiple memory layers (hops). In each layer, the imprint descriptors $\boldsymbol\pi_{\bf i}$ or $\boldsymbol\mu_{\bf i}$ from video imprint are first embedded to the output vector space and memory vector space with embedding matrices ${\bf B}$ and ${\bf M}$, respectively,
\begin{equation}
\begin{aligned}
{\rm TCG:}\ \ &{\boldsymbol b}_{\bf i} = {\bf B} \boldsymbol\pi_{\bf i},~~{\boldsymbol m}_{\bf i} = {\bf M} \boldsymbol\pi_{\bf i},\\
{\rm Epitome:}\ \ &{\boldsymbol b}_{\bf i} = {\bf B} \boldsymbol\mu_{\bf i},~~{\boldsymbol m}_{\bf i} = {\bf M} \boldsymbol\mu_{\bf i},
\end{aligned}
\end{equation}
where ${\boldsymbol b}_{\bf i}$ denotes the output vector and ${\boldsymbol m}_{\bf i}$ denotes memory vector. The memory vector ${\boldsymbol m}_{\bf i}$ is introduced to compute the weights map ${\bf P}=\{p_{\bf i}|{\bf i} \in {\bf E}\}$ with the internal state ${\boldsymbol u}$,
\begin{equation}
p_{\bf i} = {\rm avgpooling}\left({\rm softmax}\left( {\boldsymbol u}^{\top} {\boldsymbol m}_{\bf i} \right) \right).
\end{equation}
The average pooling is performed with $3\times3$ windows, stride $1$. The output vector $o$ is then computed by a weighted sum over the output vectors $b_{\bf i}$, {\it i.e.},
\begin{equation}
{\boldsymbol o}=\sum_{\bf i}{p_{\bf i}{\boldsymbol b}_{\bf i}}.
\end{equation}
For the internal state vector ${\boldsymbol u}$, the initial ${\boldsymbol u}^1$ is obtained with Equation (\ref{FA}), and the ${\boldsymbol u}^{k+1}$ in ${k+1}$ layer can be computed by
\begin{equation}
{\boldsymbol u}^{k+1}={\boldsymbol u}^{k}+{\boldsymbol o}^k.
\end{equation}

\myviolet{The obtained output vector is fed into a decision network for event categorization. The decision network can consist of only a single softmax layer or multiple fully connected layers. The recounting heat map\footnote{\myviolet{More sophisticated recounting inferences can be implemented by computing conditional heat maps based on individual memory layers as we trace the reasoning engine through the layers.}} (posterior probabilities $q(l^t={\bf i})$) of each frame shown in Figure \ref{figure_spatialMNN} is generated via the sum of all weights maps, ${\bf P}^{\rm sum}=\sum_{k}{{\bf P}^k}$.} We use ${\bf P}^{\rm sum}_{{\bf W}_{\bf i}}$ to denote the weights map cropped from ${\bf P}^{\rm sum}$ in the window ${\bf W}_{\bf i}$. Then the recounting map ${\bf R}^t$ of frame $t$ is
\begin{equation}
{\bf R}^t=\sum_{\bf i \in E}{q(l^t={\bf i}){\bf P}^{\rm sum}_{{\bf W}_{\bf i}}}.
\end{equation}
The importance score of each frame is obtained with the sum of the recounting map.

\section{Experiments}\label{Experiments}

\subsection{Datasets and evaluation protocol}

In terms of event retrieval, we validated our method on the large-scale benchmark EVVE dataset ~\cite{revaud2013event}. It contains $2,995$ videos ($620$ videos are set as queries) related to 13 specific event classes. Given a single video of an event, the task is to retrieve videos related to the same event from the dataset. The methods are evaluated based on the mean AP (mAP) computed per event. The overall performance is evaluated by averaging the mAPs over the 13 events. In addition, a large distractor dataset ($100,000$ vedios) is also provided to evaluate the retrieval performance on large-scale data.

To evaluate the event recognition and recounting, we used three datasets: EVVE, Columbia Consumer Videos (CCV) \cite{jiang2011consumer} and TRECVID MEDTest 14 (MED14) \cite{over2014trecvid}.

\myviolet{In addition, we also configured the EVVE as a small recognition dataset with 13 events}. For each event, we set the query video as the test data ($620$ videos), and treat the ground truth in the dataset as the training data. We report the top-1 classification accuracy for performance evaluation.

The Columbia Consumer Videos (CCV) dataset \cite{jiang2011consumer} contains $9,317$ YouTube videos {of} $20$ classes. We follow the protocol defined in \cite{jiang2011consumer}, {with} a training set of $4,659$ videos and a test set of $4,658$ videos. The TRECVID MEDTest 14 (MED14) \cite{over2014trecvid} is one of the most challenging datasets for event recognition containing $20$ complex events. In the training section, there are $100$ positive exemplars per event, and all events share negative exemplars with about $5,000$ videos. The test {section contains} approximately $23,000$ videos. 

For these two datasets, mAP is used as the {evaluation metric of event recognition} according to the NIST standard \cite{over2014trecvid}. \myviolet{Since no ground truth is available for the recounting task, we only provide a user study and qualitative analysis for such results.}

\begin{figure}[t]
\begin{center}
   \includegraphics[width=0.99\linewidth]{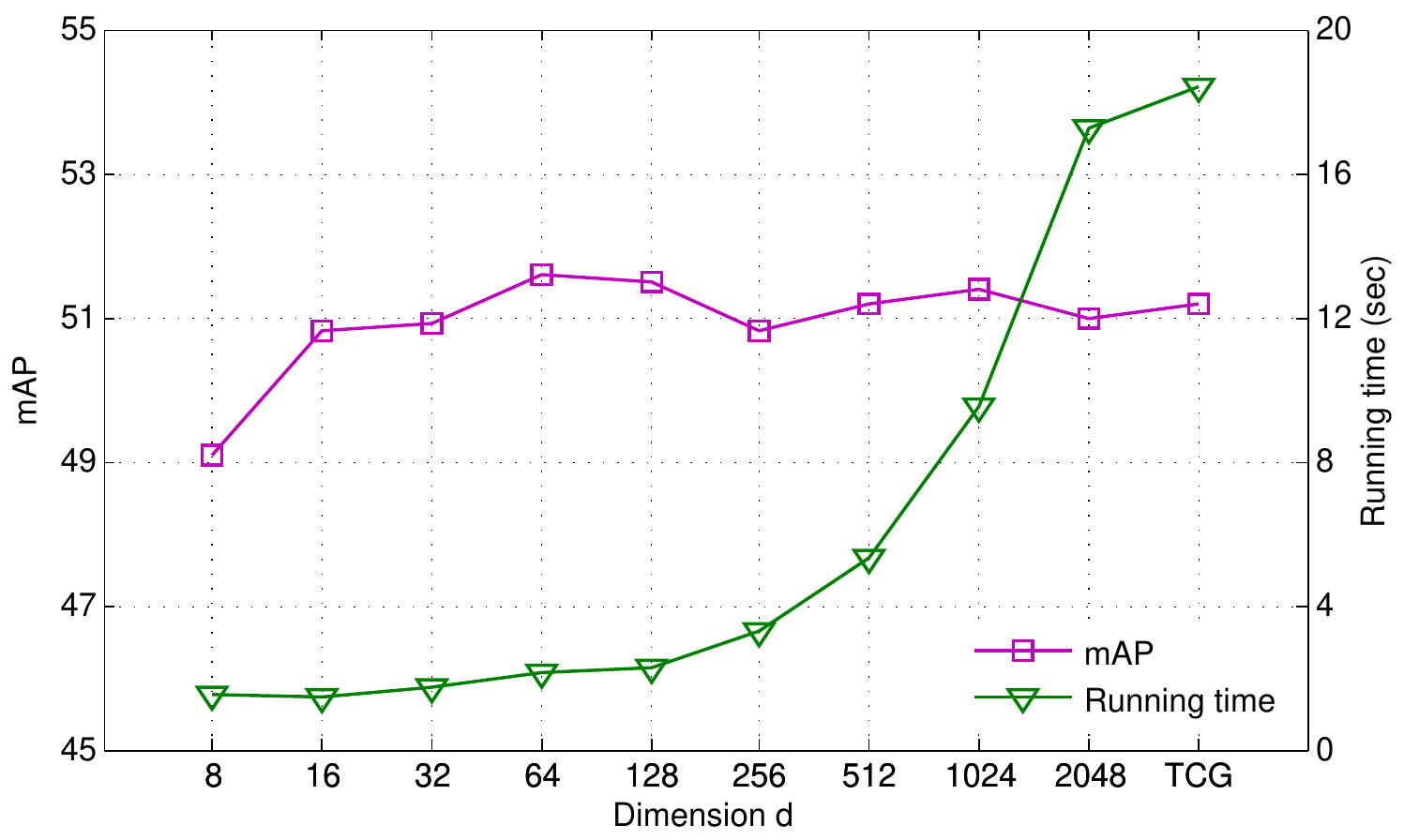}
\end{center}
   \caption{The average running time of learning epitomes for EVVE dataset with different $d$ and the corresponding retrieval performance (mAP).}
\label{figure_mAPandTime}
\end{figure}

\begin{figure}
\begin{center}
   \includegraphics[width=0.99\linewidth]{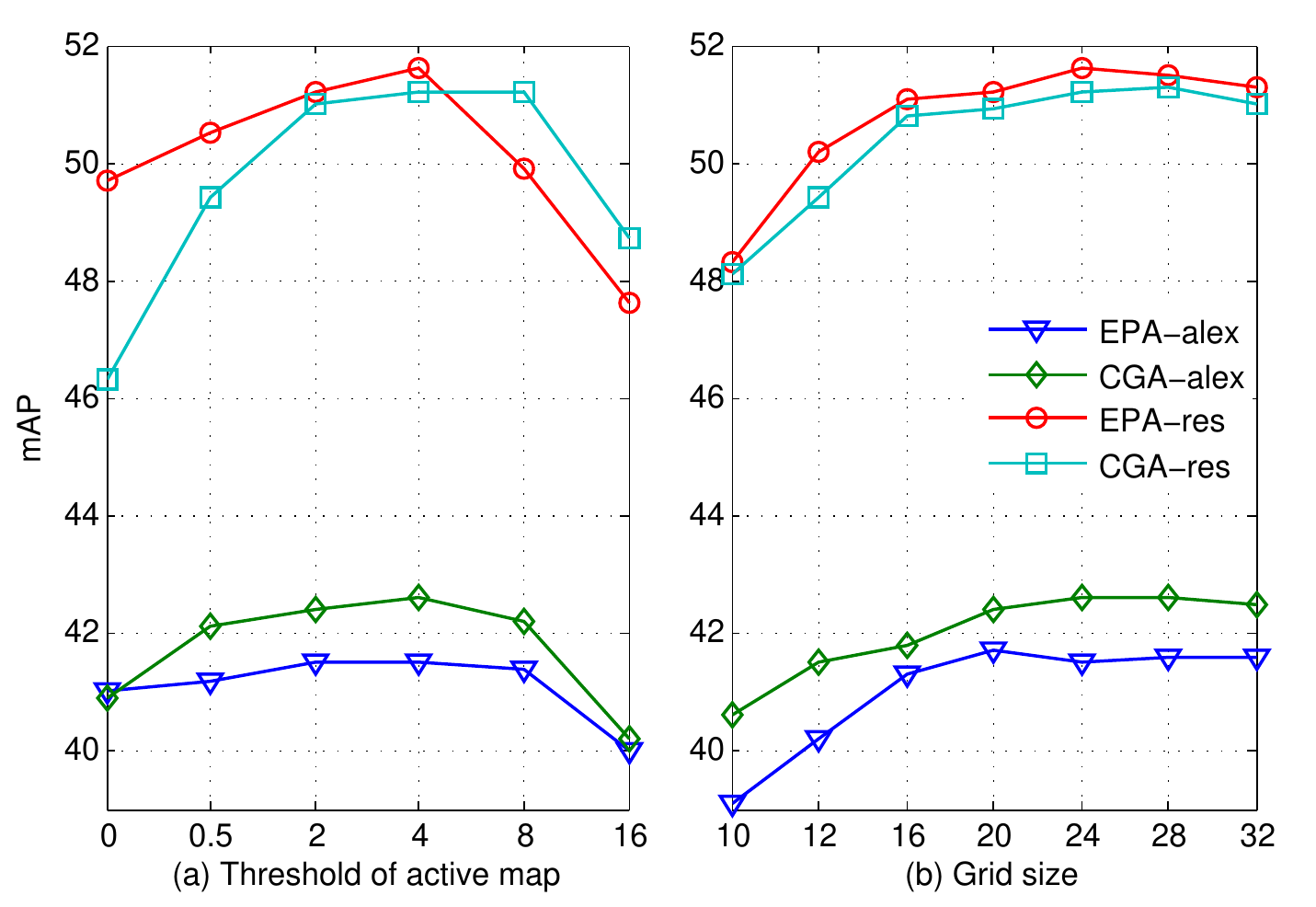}
\end{center}
   \caption{\myviolet{The influence of different (a) threshold $\tau$ of active map and (b) grid sizes {\bf E} for Counting grid aggregation (CGA) and Epitome aggregation (EPA). {alex} and {res} denote two CNN models, {\em i.e.}, AlexNet and ResNet.}}
\label{figure_cgsize}
\end{figure}

\subsection{Implementation details}

{\bf Frame-level descriptor}.
\myviolet{Given an input video, we sample $5$ frames per second ($5$ fps) to extract the CNN features. We explore various pre-trained CNN models, including AlexNet~\cite{krizhevsky2012imagenet}, VGG \cite{simonyan2014very} and ResNet-50 \cite{he2016resnet}.} We adopt the output from the last convolutional layer (after ReLU) of these models as the frame descriptors. The CNN feature maps are down-sampled to $4\times4$ with linear interpolation to fit the TCG (we set ${\bf S=}4\times4$ in TCG for {computational} efficiency). For the epitome model, the feature maps are sampled to $8\times8$ to fit the window size ${\bf W}$. In addition, we also average all the frame descriptors over the video (sum-aggregation), as the baseline to evaluate our framework.

{\bf{Post-processing.}}
For the baseline video representation, we apply the same post-processing strategy as in~\cite{babenko2015aggregating,gao2016democratic}, {\em i.e.}, the representation vector of a video is first $l_2$-normalized, and then whitened using PCA~\cite{jegou2012negative} and $l_2$-normalized again. For the imprint descriptors on the video imprint, power normalization ($\alpha=0.2$) shows better results than $l_2$-normalization in our experiments. Therefore, after feature alignment, the imprint descriptors are first power normalized, then PCA-whitened and $l_2$-normalized.

 {\bf{Re-ranking methods for event retrieval.}}
 For the event retrieval task on the {EVVE} dataset, we also employ two variants of query expansion method presented by Douze {\em et al.} \cite{douze2013stable}: Average Query Expansion (AQE) and Difference of Neighborhood (DoN). In our experiments, we set $N_1=10$ for AQE and $N_1=10,N_2=2000$ for DoN.

\setlength{\tabcolsep}{25.7pt}
\begin{table}[t]
\begin{center}
\caption{Comparison with sum-aggregation on EVVE dataset. {Sum-}, {CGA-} and {EPA-} denote sum-aggregation, counting grid aggregation, epitome aggregation, respectively. {alex} and {res} denote two CNN models, AlexNet and ResNet-50. For ResNet based representation, the vectors dimension are reduced to 1024 with PCA-whitening. {(alex+res)} denotes the concatenated vector.}

\label{table_sum_agg}
\begin{tabular}{lcc}
\toprule
Representation        & Dim.             & mAP            \\
\midrule
Sum-alex              & 256              & 38.3           \\
Sum-res               & 1024             & 46.6           \\
Sum-(alex+res)        & 1280             & 47.3           \\
\midrule
CGA-alex              & 256              & 42.6           \\
CGA-res               & 1024             & 51.2           \\
CGA-(alex+res)        & 1280             & {\bf 52.3}     \\
\midrule
EPA-alex              & 256              & 41.5           \\
EPA-res               & 1024             & 51.6           \\
EPA-(alex+res)        & 1280             & 52.1           \\
\bottomrule
\end{tabular}

\end{center}
\end{table}

{\bf{Training details for the reasoning network.}}
The reasoning network (RNet) is trained with stochastic gradient descent (SGD). The initial learning rate is $\beta = 0.025$, which is then annealed every 5 epochs by $\beta/2$ until 20 epochs are finished. All weights are initialized randomly from a Gaussian distribution with zero mean and $\sigma = 0.05$. The weights are shared among different memory layers. The batch size is 128 and the gradients with an $l_2$ norm larger than 20 are rescaled to norm 20 during the training step.

\begin{figure*}
\begin{center}
   \includegraphics[width=0.9\linewidth]{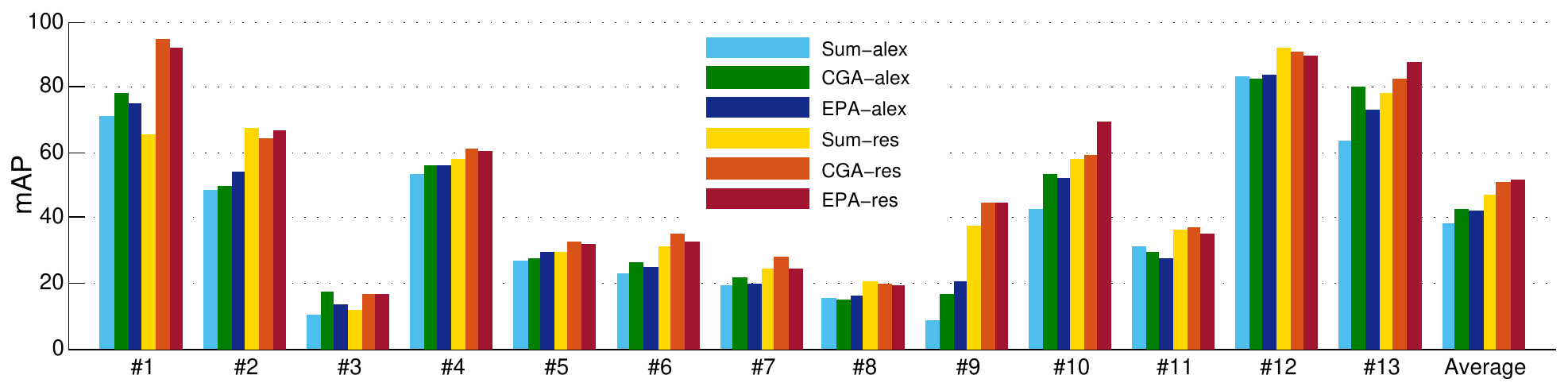}
\end{center}
   \caption{Retrieval performance (mAP) per event.}
\label{figure_mAP13classes}
\end{figure*}

\setlength{\tabcolsep}{12.5pt}
\begin{table*}[t]
\begin{center}
\caption{\myviolet{Retrieval performance compared with other methods. AQE and DoN denote the two Re-ranking methods.}}

\label{table_qe}
\begin{tabular}{lccccccc}
\toprule
                                            &      &               \multicolumn{3}{c}{EVVE}        &         \multicolumn{3}{c}{EVVE+100K}          \\
\cmidrule(lr){3-5} \cmidrule(lr){6-8}
Method                                      &Dim.  & no QE          & AQE          & DoN            & no QE          & AQE          & DoN            \\ 
\midrule
MMV                 \cite{revaud2013event}  &512   & 33.4           &  --          &  --            & 22.0           &  --          &  --            \\
CTE                 \cite{revaud2013event}  &--    & 35.2           &  --          &  --            & 20.2           &  --          &  --            \\
MMV+CTE             \cite{revaud2013event}  &--    & 37.6           &  --          &  --            & 25.4           &  --          &  --            \\
 \myviolet{TMK             \cite{poullot2015temporal}}  & \myviolet{66560} &  \myviolet{33.5}           &  \myviolet{36.1}         &  \myviolet{41.3}           &  \myviolet{25.4}           &   \myviolet{--}          &  \myviolet{34.7} 
\\
 \myviolet{SHP                 \cite{douze2013stable}}  & \myviolet{16384} &  \myviolet{36.3}           &  \myviolet{38.9}         &  \myviolet{44.0}           &  \myviolet{26.5}           &  \myviolet{30.1}         &  \myviolet{33.1}   
\\
\midrule
 \myviolet{CTE-(alex+res)}                              & \myviolet{--}    &  \myviolet{44.7}           &   \myviolet{--}          &   \myviolet{--}            &  \myviolet{40.1}           &   \myviolet{--}          &   \myviolet{--}            \\
 \myviolet{TMK-(alex+res)}                              & \myviolet{42240} &  \myviolet{46.7}           &  \myviolet{51.8}         &  \myviolet{53.0}           &  \myviolet{40.9}           &  \myviolet{48.4}         &  \myviolet{48.0}           \\
 \myviolet{SHP-(alex+res)}                              & \myviolet{40960} &  \myviolet{48.7}           &  \myviolet{54.0}         &  \myviolet{55.4}           &  \myviolet{41.5}           &  \myviolet{48.7}         &  \myviolet{48.9}           \\
Sum-(alex+res)                              &1280  & 47.3           & 53.1         & 55.2           & 38.7           & 45.8         & 47.1           \\
\midrule
CGA-(alex+res)                              &1280  & {\bf 52.3}     & 58.5         & {\bf 60.1}     & 42.9           & 50.4         & 52.7           \\
EPA-(alex+res)                              &1280  & 52.1           & {\bf 59.3}   & 59.9           & {\bf 43.9}     & {\bf 53.2}   & {\bf 53.8}     \\
\bottomrule
\end{tabular}
\end{center}
\end{table*}

\subsection{Complexity analysis}
The most time consuming step is constructing video imprint for input videos. As shown in Figure \ref{figure_mAPandTime}, the average running time of TCG (with ResNet features) for EVVE (about 1200 frames per video) implemented on the GPU platform (K40 with MATLAB parallel computing toolbox) is about 18 seconds. As discussed in \cite{perina2011image,perina2015capturing}, with efficient use of cumulative sums, the computational complexity of learning TCG with the EM algorithm grows at most linearly with the product of counting grid size and the tessellation sections $E_x \cdot E_y \cdot S_x \cdot S_y$. 

For the epitome, the computational complexity without the efficient two-step scheme for  {$D$-dimensional} input feature maps and EM iteration times $n$ ($n \approx 30$) is $O(nD)$. With the proposed efficient two-step scheme, the computational complexity becomes $O(nd+D)$. When $d \ll D$, the efficient two-step scheme can significantly accelerate the learning stage. Figure \ref{figure_mAPandTime} shows the average running time of learning epitomes for the EVVE dataset with different $d$ and the corresponding retrieval performance (mAP). When $d=64$, without any performance {losses}, the efficient two-step scheme achieves almost an order of magnitude speed-up. 

 \begin{figure*}[t]
\begin{center}
   \includegraphics[width=0.95\linewidth]{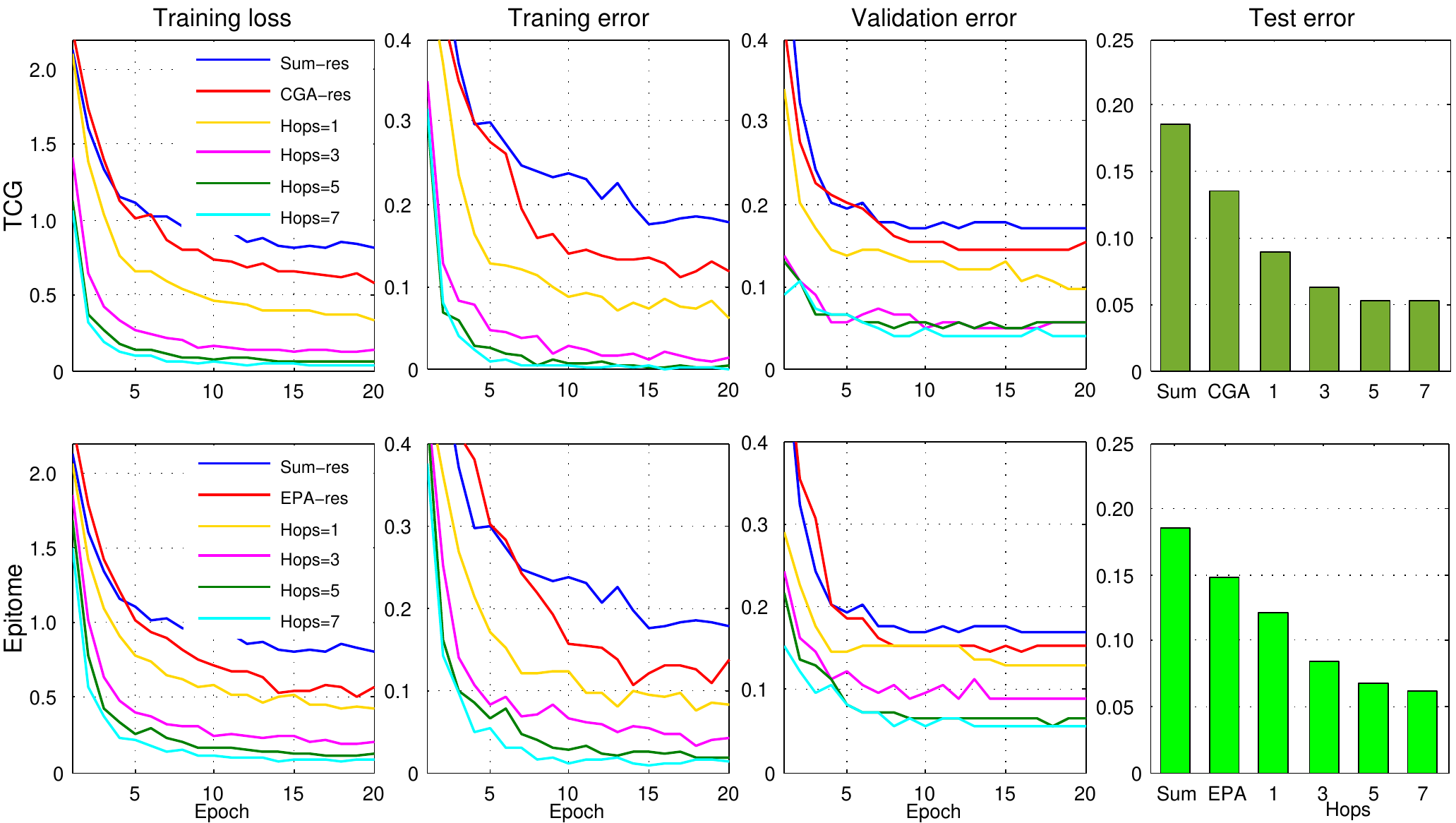}
\end{center}
   \caption{The influence of RNet with increased hops on EVVE dataset. Best viewed in color.}
\label{figure_EVVE}
\end{figure*}

\subsection{Evaluation results on event retrieval}

\subsubsection{Parameter analysis}

{\bf{Threshold} $\tau$ \bf{of the active map.}} Figure~\ref{figure_cgsize} (a) shows the retrieval performance with different threshold $\tau$ used for the active map construction. We can observe that increasing $\tau$ helps filter out some very short shots (with {small number of} frames) which are usually {irrelevant}. We set $\tau=4$ in the subsequent experiments. 

{\bf{Grid size} ${\bf E}$.} To evaluate the influence of the grid size ${\bf E}$ of the TCG and epitome models, we first fix the window size (${\bf W}=8\times 8$) of TCG, epitome, and tessellation size (${\bf S}=4\times4$) of TCG. \myviolet{Subsequently, we choose $7$ different counting grid sizes to perform the feature alignment with the epitome and TCG model, respectively. The performance with regard to size choices is presented in Figure~\ref{figure_cgsize} (b). No further improvement can be obtained when $E_x = E_y >24$. Therefore, the size of the grid is fixed at $24$ for {all} following experiments.}

\subsubsection{Comparison with sum-aggregation}
We refer to our unsupervised flow (combining the feature alignment and aggregation steps) on ER3 as counting grid aggregation (CGA) for TCG based video imprint and epitome aggregation (EPA) for epitome based video imprint. Table~\ref{table_sum_agg} shows the retrieval performance compared with baselines. We evaluate the CGA and EPA on two different CNN models, AlexNet \cite{krizhevsky2012imagenet} and ResNet-50 \cite{he2016resnet}. Figure \ref{figure_mAP13classes} shows the retrieval performance for each event class of EVVE dataset. The IDs of events are the same with \cite{revaud2013event}. We can see that the video representation based on ResNet-50 demonstrates superior performance in most event categories compared with AlexNet. And the proposed aggregation methods can further boost the retrieval performance.

As shown in Table~\ref{table_sum_agg}, compared with Sum-aggregation, CGA and EPA both obtain better retrieval performance with the benefits from feature alignment step that can suppress the redundancy among frames. In addition, consistent improvement can be observed for different CNN models (AlexNet and ResNet-50). After concatenating the video representation vectors from these two CNN models, the retrieval performance can be further improved, ${\rm mAP=52.3}$ for CGA and ${\rm mAP=52.1}$ for EPA.

\subsubsection{Comparison with state-of-the-arts}

\myviolet{ In Table \ref{table_qe}, we can see that the sum-aggregation with CNN features already outperforms previous work~\cite{revaud2013event,douze2013stable,poullot2015temporal}. After merging with 100K distractors, the mAPs of CGA-(alex+res) and EPA-(alex+res) achieve $42.9$ and $43.9$ respectively, which are still better than the baseline ($\rm mAP=38.7$) and previous work~\cite{revaud2013event,douze2013stable,poullot2015temporal}. For fair comparison, we also reimplement previous aggregation methods (CTE, TMK and SHP) based on the same CNN features with CGA and EPA. Note that MMV is in fact the sum-aggregation based on the multi-VLAD frame feature \cite{revaud2013event}, which employs the same aggregation method with our baseline method. }
 
\myviolet{ As shown in Table \ref{table_qe}, frame-level descriptors have a huge impact on performance. {CNN feature-based aggregation methods} achieve better retrieval performance than {their original handcrafted feature-based implementations}. However, based on the same frame-level descriptors, our aggregation methods still obtain better results compared with existing methods. In addition, previous aggregation methods usually lead to much higher dimension in the video representation (which may reduce the efficiency of the retrieval task). Based on the initial retrieval results, the query expansion can further boost the performance. We achieve  $8.5\%$ and $8.9\%$ improvement compared with previous result ($\rm mAP=55.4$ for SHP~\cite{douze2013stable} combined with CNN feature) and the baseline ($\rm mAP=55.2$) on the EVVE dataset, respectively. Consistent improvement is also observed with query expansion on the large dataset (EVVE+100K). }

\subsection{Evaluation results on event recognition}

\subsubsection{Parameter analysis}

{\bf{Structure of the reasoning network.}} For the EVVE dataset, we set a softmax layer as the decision network. The video imprint is generated based on the ResNet-50 \cite{he2016resnet} model and its imprint descriptors are first reduced to $256$ dimension with PCA-whitening before being fed to the reasoning network (RNet). For the CCV and MED14 datasets, we add a fully connected layer in front of the softmax layer as the decision network for better performance. Besides the ResNet-50 model, we also evaluate the framework with VGG (16 layers) \cite{simonyan2014very} model on these two datasets. The dimension of the imprint descriptors is set to 1024 and 512 for ResNet-50 and VGG, respectively. For all the datasets, the internal vectors ${\boldsymbol b}_{\bf i}$ and ${\boldsymbol m}_{\bf i}$ have the same dimensions with the input imprint descriptors.

\setlength{\tabcolsep}{14pt}
\begin{table}[t]
\begin{center}
\caption{Comparison with sum-aggregation and CGA/EPA. {Sum-}, {CGA-} and {EPA-} denote sum-aggregation, counting grid aggregation and epitome aggregation, respectively. RNet- denote the reasoning network. {vgg} and {res} denote two CNN model, VGG and ResNet-50.  {(vgg+res)} denotes the later fusion result.}

\label{table_sumCGA}
\begin{tabular}{ccccc}
\toprule
                                      &Method        & vgg        & res        & (vgg+res)  \\
\midrule
\multirow{5}{*}{\rotatebox{90}{CCV}}  &Sum-          & 74.3       & 75.3       & 78.1       \\
                                      &CGA-          & 75.7       & 76.6       & 79.1       \\
                                      &EPA-          & 75.2       & 76.7       & 78.3       \\
                                      &CG-RNet-      & 76.7       & 78.5       & {\bf 79.9} \\
                                      &EP-RNet-      & 75.8       & 78.2       & 79.0       \\
\midrule
\multirow{5}{*}{\rotatebox{90}{MED14}}&Sum-          & 26.0       & 30.4       & 32.8       \\
                                      &CGA-          & 30.5       & 32.2       & 33.7       \\
                                      &EPA-          & 31.2       & 32.4       & 34.4       \\
                                      &CG-RNet-      & 32.8       & 34.2       & {\bf 36.9} \\
                                      &EP-RNet-      & 32.6       & 34.3       & 36.3       \\
\bottomrule
\end{tabular}

\end{center}
\end{table}

\setlength{\tabcolsep}{11.7pt}
\begin{table}[t]
\begin{center}
\caption{Comparison with other methods. MA+CG/EP-RNet-(vgg+res) denote the results fused with audio and motion information using adaptive fusion method \cite{wu2016multi}. IDT+CG/EP-RNet-(vgg+res) denote the results fused with improved dense trajectories \cite{wang2013action}.}

\label{table_other}
\begin{tabular}{clccc}
\toprule
                                      &Method                                      & mAP        & Recounting \\
\midrule
\multirow{9}{*}{\rotatebox{90}{CCV}}  &Lai {\it et al.} \cite{lai2014video}        & 43.6       & $\surd  $  \\
                                      &Jiang {\it et al.} \cite{jiang2011consumer} & 59.5       & $\times $  \\
                                      &Wu {\it et al.} \cite{wu2014exploring}      & 70.6       & $\times $  \\
                                      &Nagel {\it et al.} \cite{nagel2015event}    & 71.7       & $\times $  \\
                                      &Wu {\it et al.} \cite{wu2016multi}          & 84.9       & $\times $  \\
                                      \cmidrule(lr){2-4}
                                      &CG-RNet-(vgg+res)                           & 79.9       & $\surd  $  \\
                                      &EP-RNet-(vgg+res)                           & 79.0       & $\surd  $  \\
                                      &MA+CG-RNet-(vgg+res)                        & {\bf 87.1} & $\surd  $  \\
                                      &MA+EP-RNet-(vgg+res)                        & 86.8       & $\surd  $  \\
\midrule
\multirow{8}{*}{\rotatebox{90}{MED14}}&IDT \cite{wang2013action,over2014trecvid}   & 27.6       & $\times $  \\
                                      &Gan {\it et al.} \cite{gan2015devnet}       & 33.3       & $\surd  $  \\
                                      &Xu {\it et al.} \cite{xu2015discriminative} & 36.8       & $\times $  \\
                                      &Zha {\it et al.} \cite{BMVC2015_60}         & 38.7       & $\times $  \\
                                      \cmidrule(lr){2-4}
                                      &CG-RNet-(vgg+res)                           & {36.9}     & $\surd  $  \\
                                      &EP-RNet-(vgg+res)                           & 36.3       & $\surd  $  \\
                                      &IDT+CG-RNet-(vgg+res)                       & {\bf 40.2} & $\surd  $  \\
                                      &IDT+EP-RNet-(vgg+res)                       & 39.8       & $\surd  $  \\
\bottomrule
\end{tabular}
\end{center}
\end{table}

{\bf{Number of memory layers.}} Figure \ref{figure_EVVE} illustrates the influence of RNet with increased hops on the EVVE dataset. To make a fair comparison, we employ the same decision network as classifier for the baselines and the output from RNet. We compare with two representations, one is the video representation with sum-aggregation, and the other is either the counting grid aggregation (CGA) or the epitome aggregation (EPA) which depends on the feature alignment step. In fact, if we fix the value of the weights map equal to the active map, the RNet will reduce to the CGA or EPA, {\it i.e.}, the unsupervised flow in Figure \ref{framework}.  We can see that CGA/EPA provides better performance than sum-aggregation, and the RNet can further refine the video representation and lead to better recognition accuracy than the two baselines. In addition, the gain is also increased with more hops. Consistent gains are observed on both CCV and MED14 datasets. We set the ${\rm hops}=3$ in the following experiments on the CCV and MED14 datasets.

\begin{figure}[t]
\begin{center}
   \includegraphics[width=0.99\linewidth]{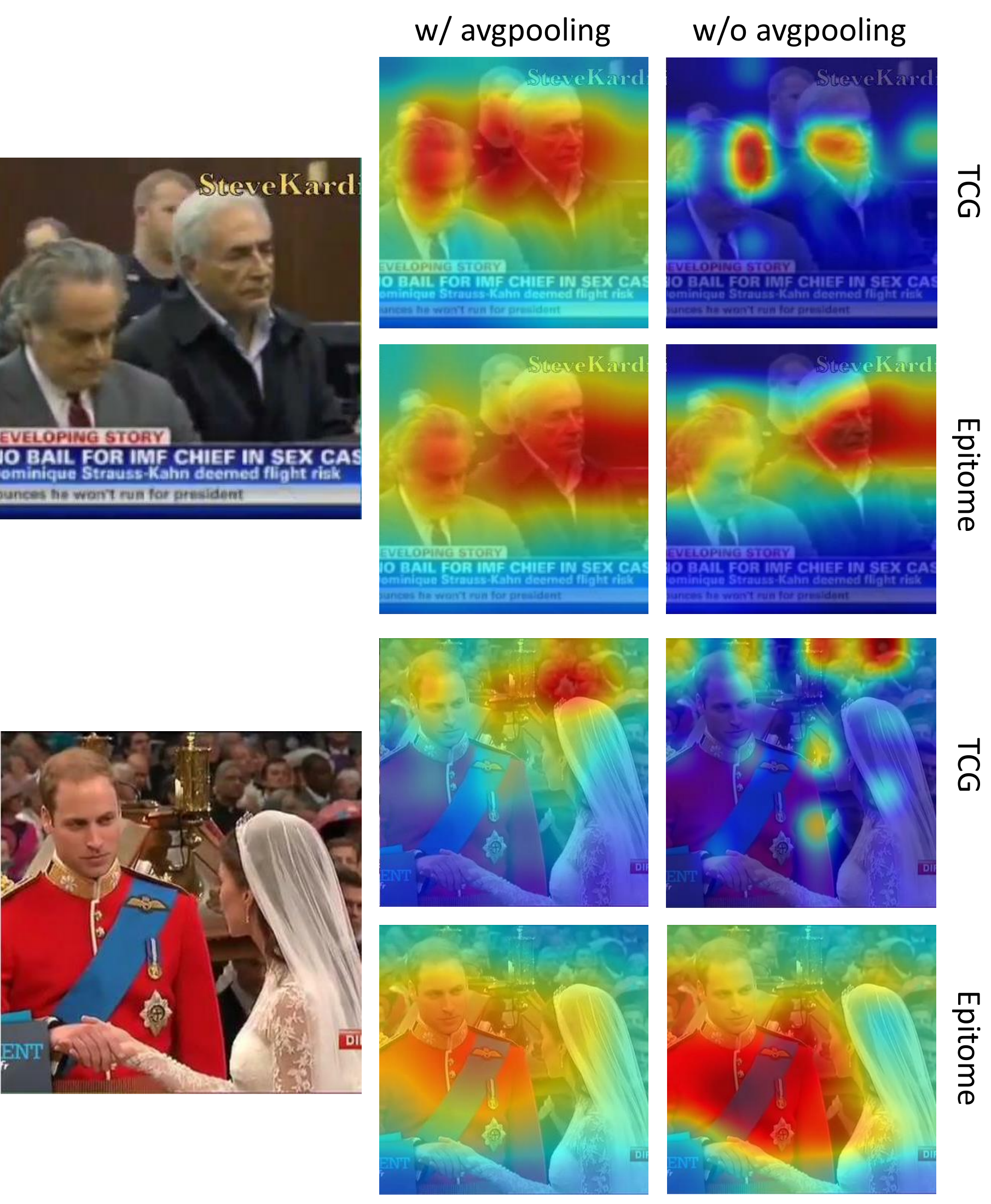}
\end{center}
   \caption{Influence of the average pooling layer in RNet. The middle column shows the recounting map of RNet. The right column shows the recounting map with avg-pooling layer removed. Best viewed in color.}
\label{figure_avg}
\end{figure}

\subsubsection{Performance on CCV and MED14}

Table \ref{table_sumCGA} shows the recognition performance (mAP) of RNet and baseline methods. With the benefit from re-weighting the video imprint, the RNet achieves better results on {both} CCV (mAP = $79.9/79.0$) and MED14 (mAP = $36.9/36.3$) datasets compared with sum-aggregation and CGA/EPA. In addition, on the CCV dataset, we also employ the same strategy as \cite{wu2016multi} to combine motion and audio features with our appearance-based representation. \myviolet{As shown in Table \ref{table_other}, the fusion strategy further boosts the recognition performance (mAP = $87.1/86.8$). On MED14 dataset, the proposed method sets a new performance record (mAP = $40.2$) after fusing motion features (IDT) in a similar way to \cite{BMVC2015_60}, with the additional convenience of simultaneously providing recounting results. }

\subsection{Evaluation results on event recounting}

\begin{figure}[t]
\begin{center}
   \includegraphics[width=0.95\linewidth]{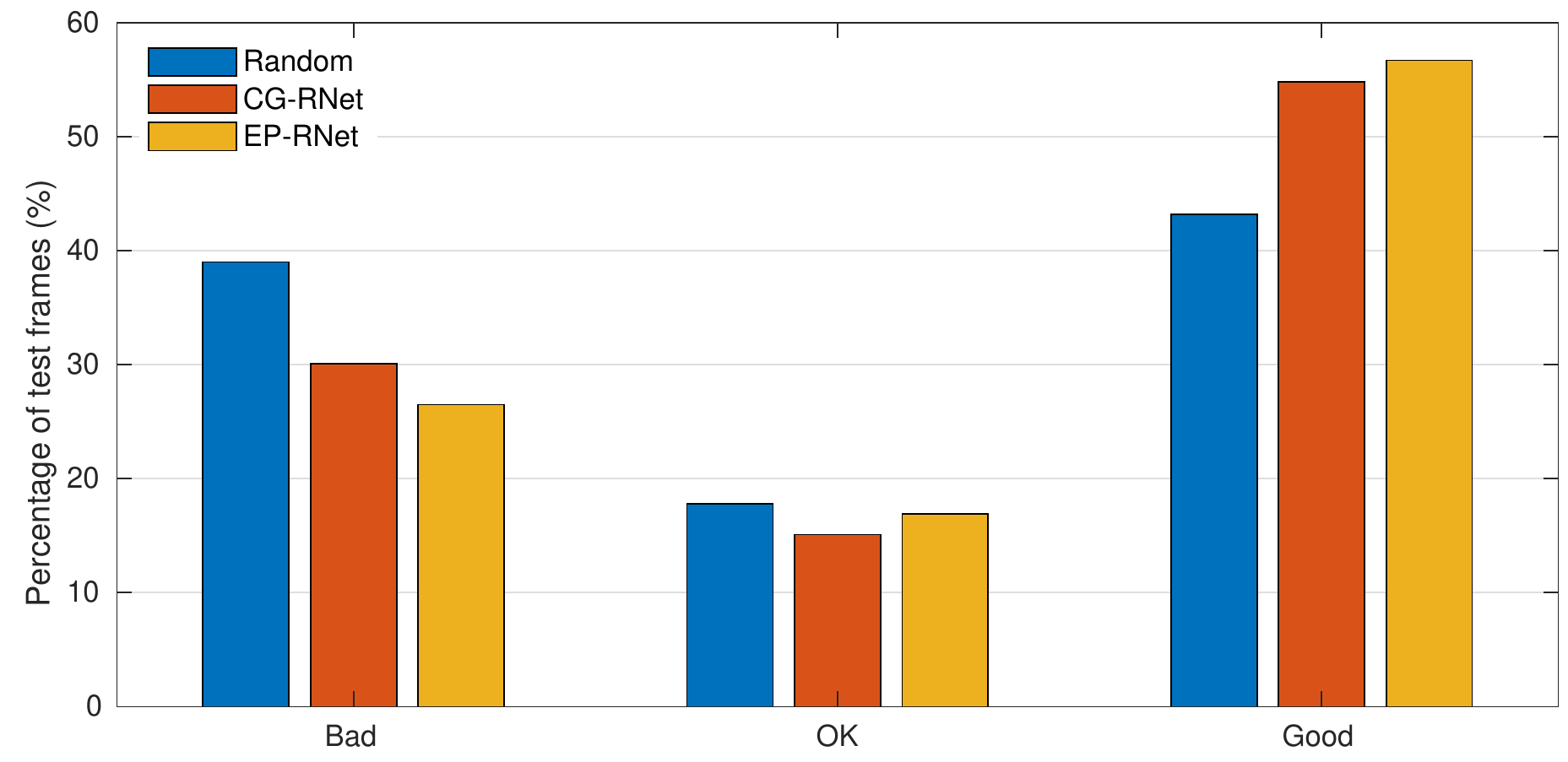}
\end{center}
   \caption{\myviolet{The statistical results of user study. The average percentage of key frames belong to {\bf Bad, OK} and {\bf Good} are shown with different color bar. Best viewed in color.}}
\label{figure_userStudy}
\end{figure}

\begin{figure}[t]
\begin{center}
   \includegraphics[width=0.99\linewidth]{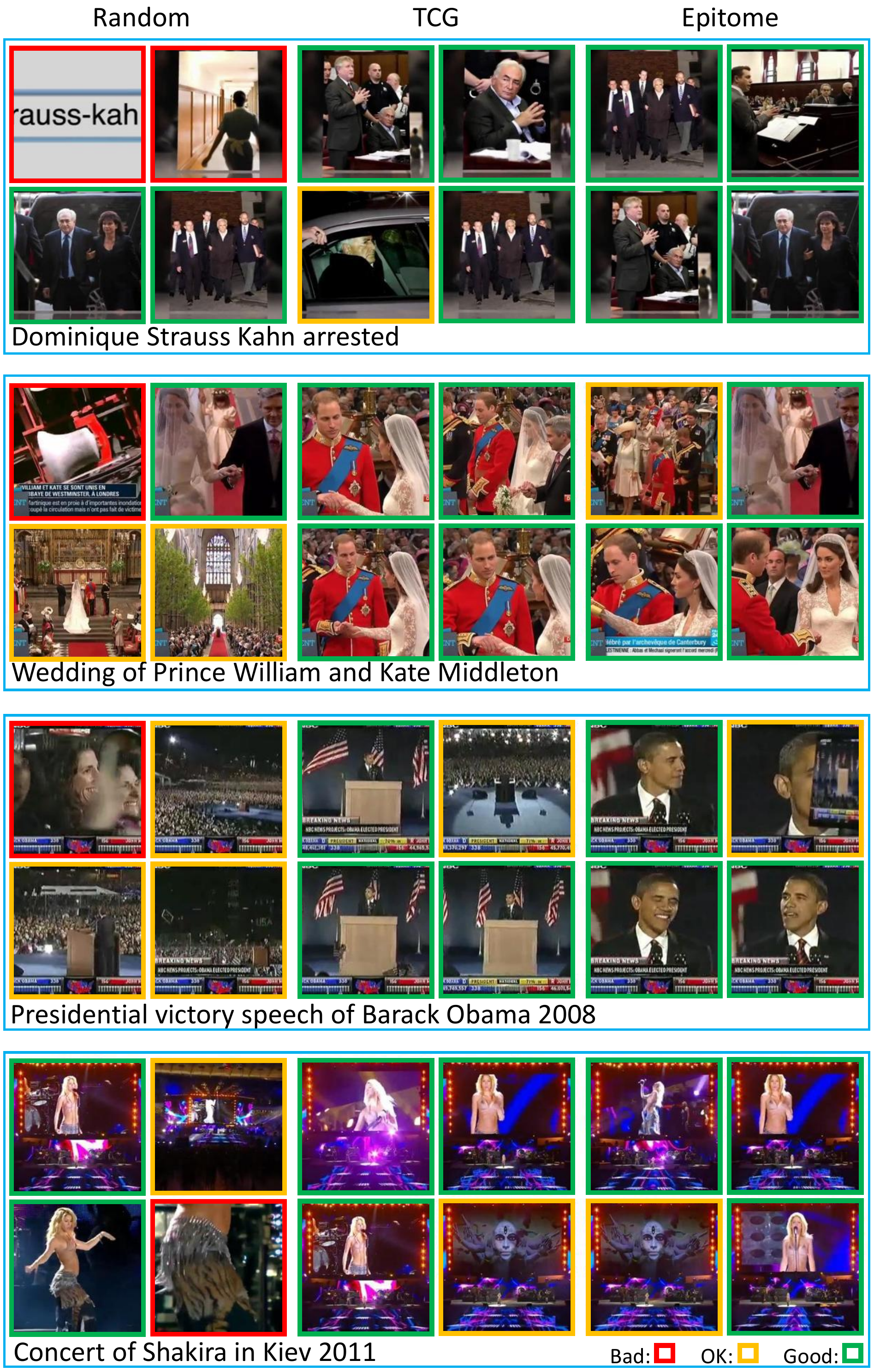}
\end{center}
   \caption{Examples of the key frames for user study. Each blue box contains the key frames proposed from a test video with three strategies, {\it i.e.}, random selection, recounting results based on TCG or epitome. Each strategy proposes four key frames to represent the video. the scores ranked by the users are shown with different color boxes, red for {\bf Bad}, yellow for {\bf OK} and green for {\bf Good}. Best viewed in color.}
\label{figure_userstudyExamples}
\end{figure}

{\bf{Influence of average pooling.}} In contrast to the original memory networks \cite{sukhbaatar2015end}, we add an average pooling layer inside the memory layer, which takes advantage of the spatial organization of the information in the video imprint. Figure \ref{figure_avg} demonstrates the influence of adding the average pooling layer. We can see that the recounting maps are smoother and more reasonable, especially for TCG based video imprint. In addition, with benefits from finer resolution of input feature maps, the epitome based video imprint can hold more spatial information, and the recounting results are more reasonable than TCG based video imprint.

{\bf{User study.}} \myviolet{The evaluation of event recounting is not easy since there is no ground truth information. To assess the quality of our recounting results, we randomly sample 200 videos from 620 test videos of the EVVE dataset for user study. First, for each test video, we sample 4 key frames either randomly or based on the importance score from the recounting map {\bf R} (3 groups in total: randomly selected, based on important scores from TCG or Epitome). Then, we invited 50 users to score the key frames with {\bf \{Bad, OK, Good\}} based on the relevance with the labeled event. Here, {\bf Bad} means that the selected frame is irrelevant with its event label, {\it i.e.}, users cannot recognize the event category by the selected frame. {\bf OK} means that users can identify the event category by the selected frame {with reasonable amount of guesswork}. {\bf Good} means that the event category can be identified {beyond doubt} by the selected frame and users are {highly} confident with the judgment.}

The statistical results are shown in Figure \ref{figure_userStudy}. It presents the average percentage of key frames belong to {\bf Bad, OK} or {\bf Good}, respectively. \myviolet{Compared with random selection, the key frames proposed with recounting map are more likely to be relevant with the corresponding event, the percentage of bad proposals is reduced from $39\%$ to $26\%$. Note that epitome based recounting results are slightly better than TCG due to richer spatial information.} Figure \ref{figure_userstudyExamples} shows the examples of key frames for user study. Each blue box contains a test video represented by 4 key frames. The key frames are proposed with three strategies, {\it i.e.}, random selection, recounting based on TCG or epitome. The scores ranked by the users are shown with different color boxes. 

\begin{figure*}[t]
\begin{center}
   \includegraphics[width=0.95\linewidth]{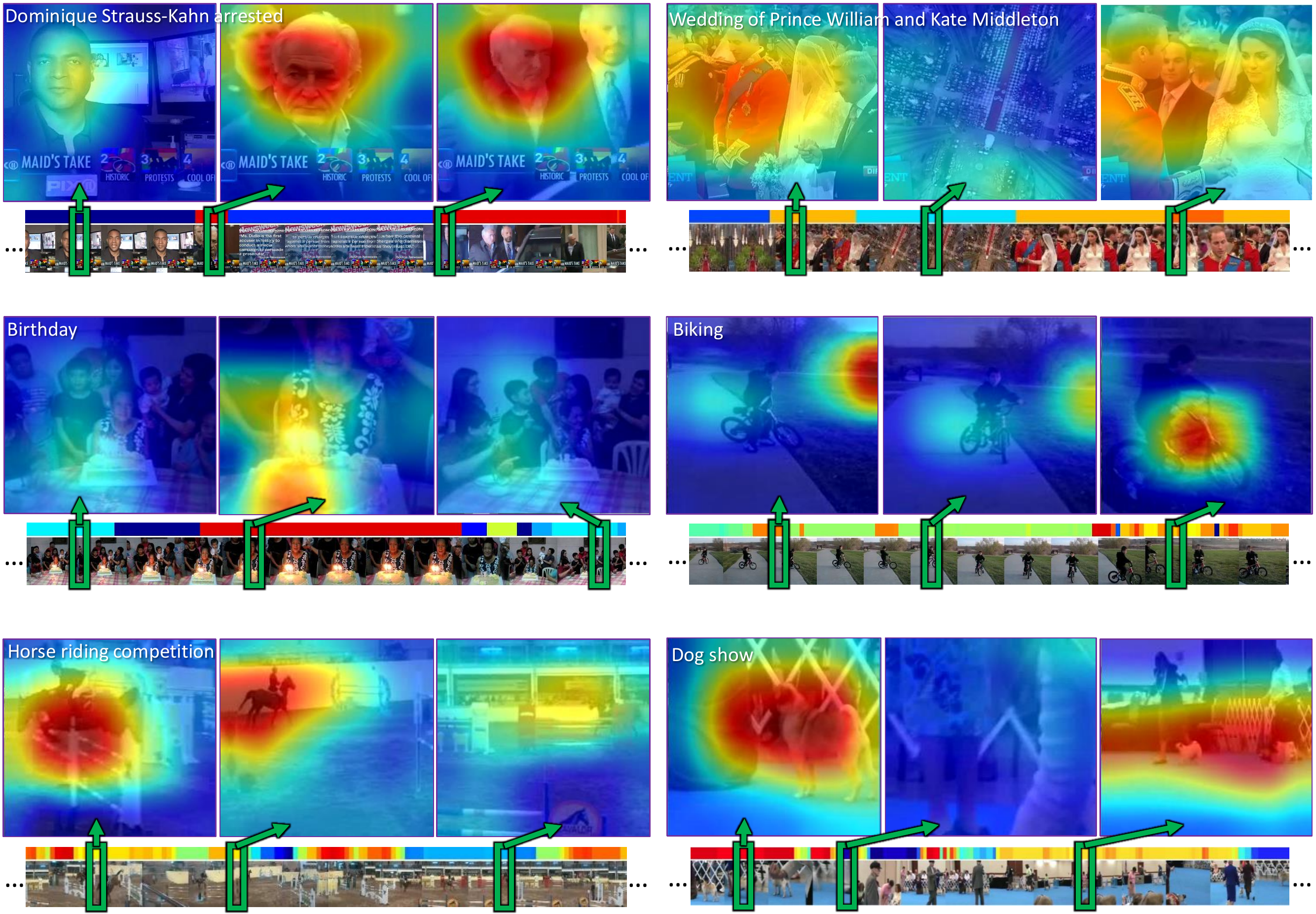}
\end{center}
   \caption{Examples of event recounting results. We use heat map to indicate the recounting map. The key areas related to the event in each frame is painted with red color. The importance score which is computed by the sum of recounting map is shown with color bar (red for important frames) upon the video frame flow. Best viewed in color.}
\label{figure_Rmap}
\end{figure*}

{\bf{Recounting map.}} Figure \ref{figure_Rmap} illustrates some examples of the recounting results. The heat map is used to visualize the recounting map (the map is rescaled to the same size with the frame). Since no ground truth recounting is available, we can only provide some examples as shown in Figure \ref{figure_Rmap}. We can see that our recounting process can not only provide the importance score of each frame, but also indicate the most relevant areas inside each frame. However, due to the coarse resolution of the input feature maps, the spatial-level recounting results are also very coarse. Nevertheless, the recounting heat map may be used as a good prior for other post-processing methods, {\it e.g.}, object segmentation.

\section{Conclusion}\label{Conclusion}


\myviolet{In this paper, we propose an efficient and effective event retrieval, recognition and recounting framework (ER3), based on our proposed video imprint representation. In the proposed framework, a dedicated feature alignment module is incorporated for redundancy removal across frames to produce the compact intermediate tensor representation, {\em i.e.,} the video imprint. Subsequently, the video imprint is processed by a reasoning network for event recognition/recounting, and by a feature aggregation module for event retrieval. Thanks to the attention mechanism inspired by the memory networks in language modeling, the proposed reasoning network is capable of simultaneous event recognition and event recounting. With the event retrieval task, the compact video representation aggregated from the video imprint contributes to better retrieval results.}
%


%



\ifCLASSOPTIONcompsoc
  \section*{Acknowledgments}
\else
  \section*{Acknowledgment}
\fi

This work was supported partly by National Key R$\&$D Program of China Grant 2017YFA0700800, National Natural Science Foundation of China Grants 61629301, 61773312, 91748208, and 61503296, China Postdoctoral Science Foundation Grants 2017T100752 and 2015M572563.

\bibliographystyle{abbrv}
\bibliography{ER3_pami}




%


%

\vspace{-40pt}

\begin{IEEEbiography}[{\includegraphics[width=1in,height=1.25in,clip,keepaspectratio]{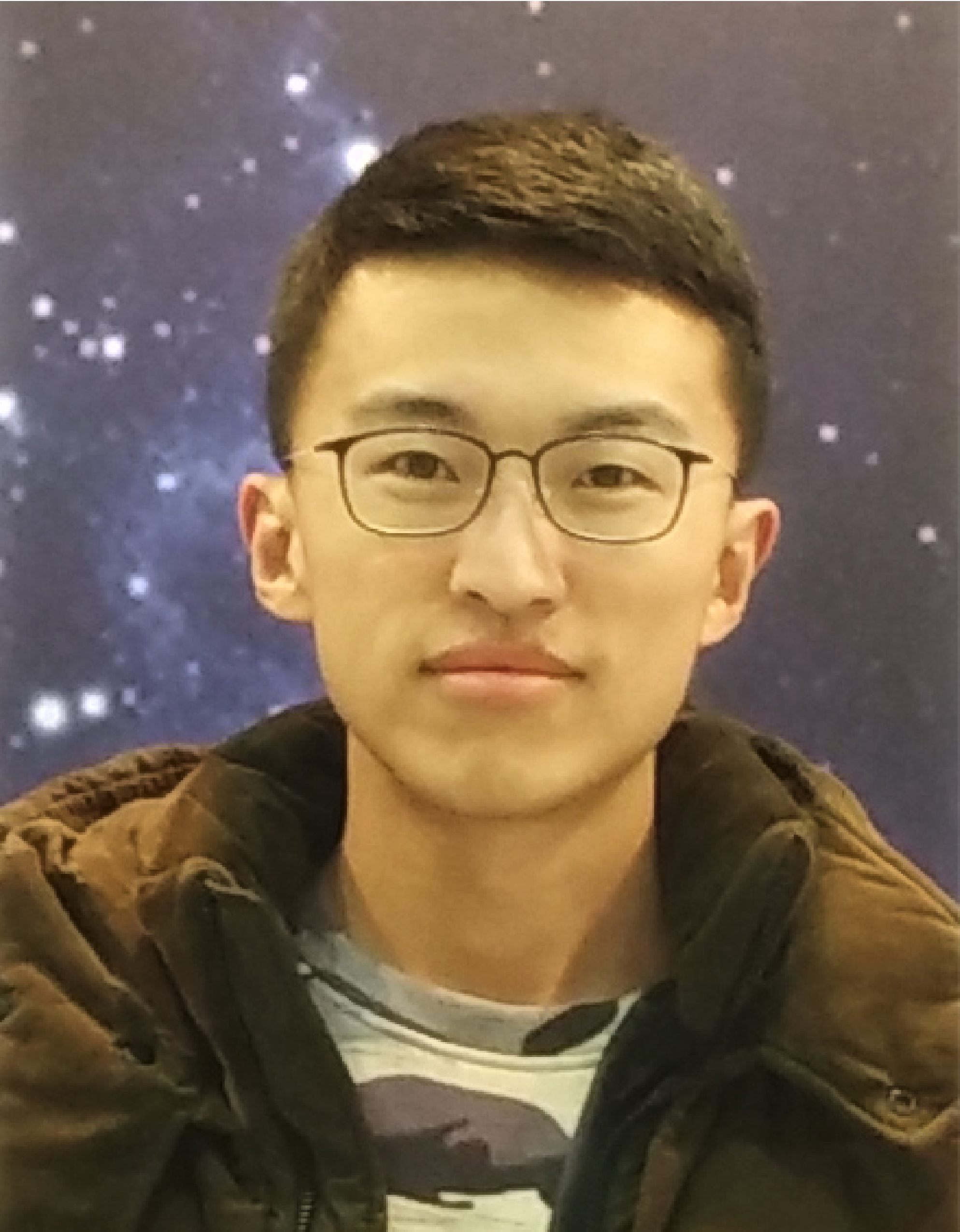}}]{Zhanning Gao}
received the B.S. degree in automatic control engineering from Xi'an Jiaotong University, Xi'an, China, in 2012. He is currently a Ph.D. candidate in Institute of Artificial Intelligence and Robtics, Xi'an Jiaotong University. He was a research intern in
Visual Computing Group in Microsoft Research Asia from 2015 to 2017. His research interests include compact image/video representation, large scale content based multimedia retrieval and complex event video analysis.
\end{IEEEbiography}

\begin{IEEEbiography}[{\includegraphics[width=1in,height=1.25in,clip,keepaspectratio]{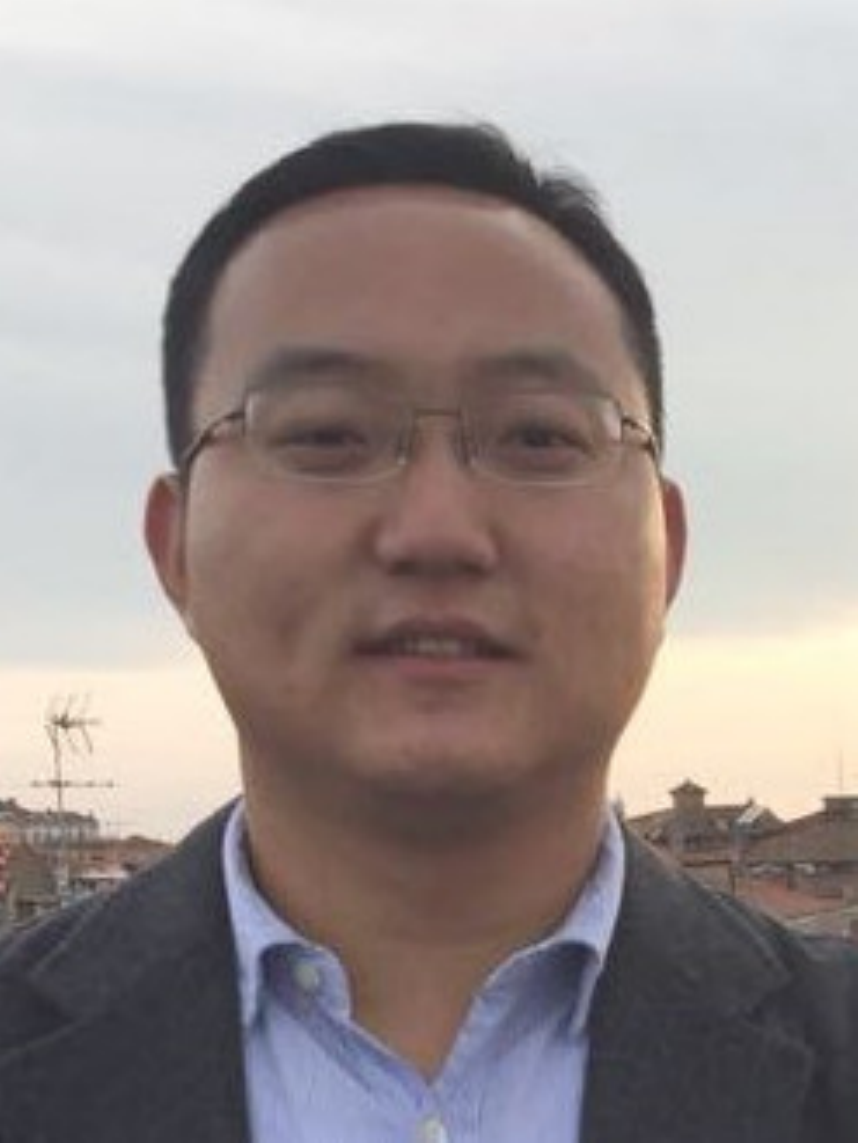}}]{Le Wang}
(M'14) received the B.S. and Ph.D. degrees in Control Science and Engineering from Xi'an Jiaotong University in 2008 and 2014, respectively. From 2013 to 2014, he was a visiting Ph.D. student with Stevens Institute of Technology. From 2016 to 2017, he is a visiting scholar with Northwestern University. He is currently an Associate Professor with the Institute of Artificial Intelligence and Robotics of Xi'an Jiaotong University. His research interests include computer vision, machine learning, and their application for web images and videos. He is the author of more than 20 peer reviewed publications in prestigious international journals and conferences. He is a member of the IEEE.
\end{IEEEbiography}

\begin{IEEEbiography}[{\includegraphics[width=1in,height=1.25in,clip,keepaspectratio]{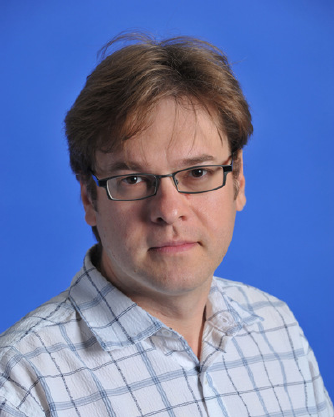}}]{Nebojsa Jojic}
received the PhD degree from the University of Illinois at Urbana-Champaign in 2001, where he received a Microsoft Fellowship in 1999 and a Robert T. Chien Excellence in Research award in 2000. He has been a researcher at Microsoft Research in Redmond, Washington, since 2000. He has published more than 100 papers in the areas of computer vision, machine learning, signal processing, computer graphics, and computational biology.
\end{IEEEbiography}

\vspace{-60pt}

\begin{IEEEbiography}[{\includegraphics[width=1in,height=1.25in,clip,keepaspectratio]{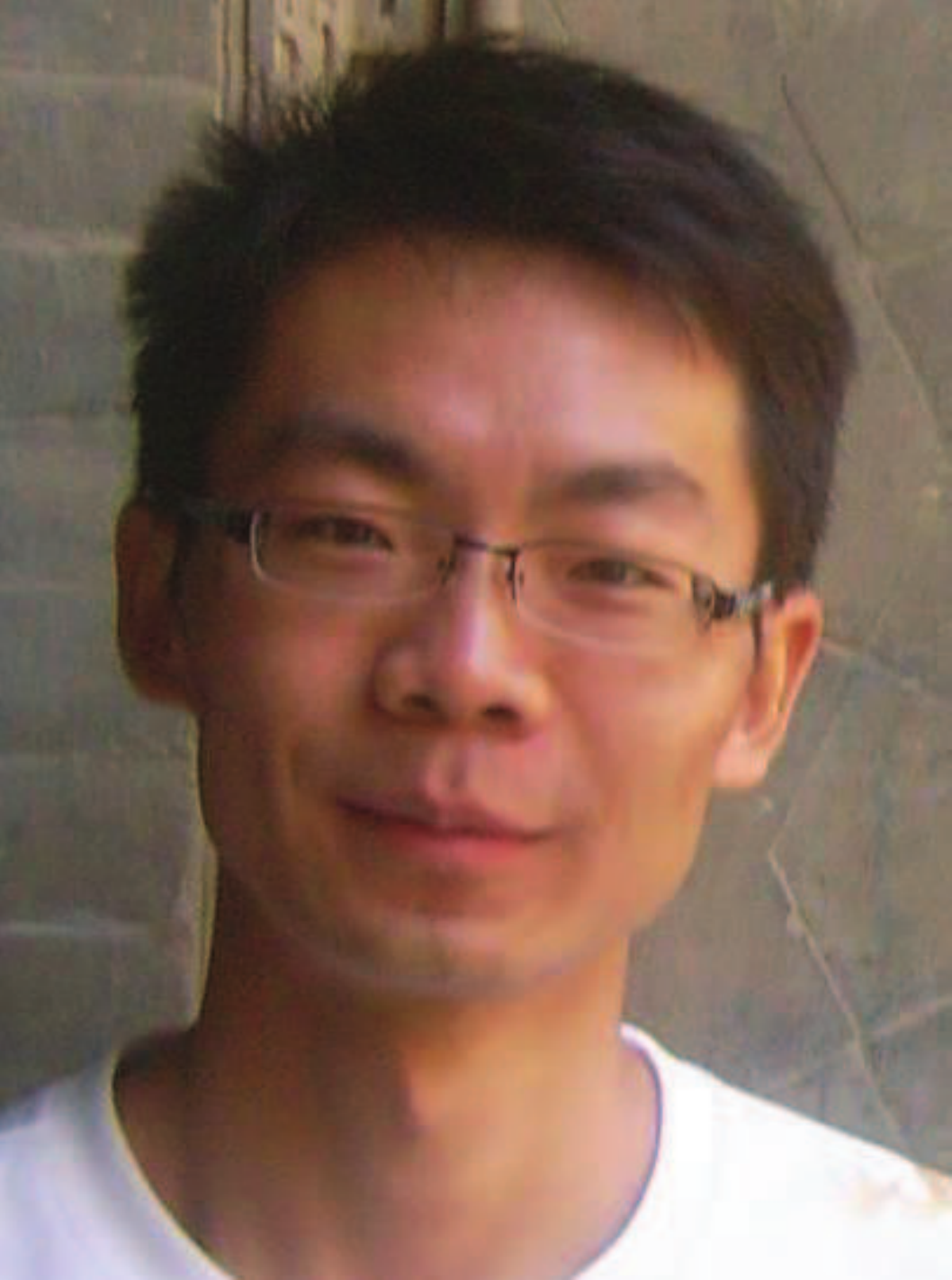}}]{Zhenxing Niu}
received the Ph.D. degree in Control Science and Engineering from Xidian University, Xi'an, China, in 2012. From 2013 to 2014, he was a visiting scholar with University of Texas at San Antonio, Texas, USA. He is a Researcher at Alibaba Group, Hangzhou, China. Before joining Alibaba Group, he is an Associate Professor of School of Electronic Engineering at Xidian University, Xi'an, China. His research interests include computer vision, machine learning, and their application in object discovery and localization. He served as PC member of CVPR, ICCV, and ACM Multimedia. He is a member of the IEEE.
\end{IEEEbiography}%

\vspace{-60pt}

\begin{IEEEbiography}[{\includegraphics[width=1in,height=1.25in,clip,keepaspectratio]{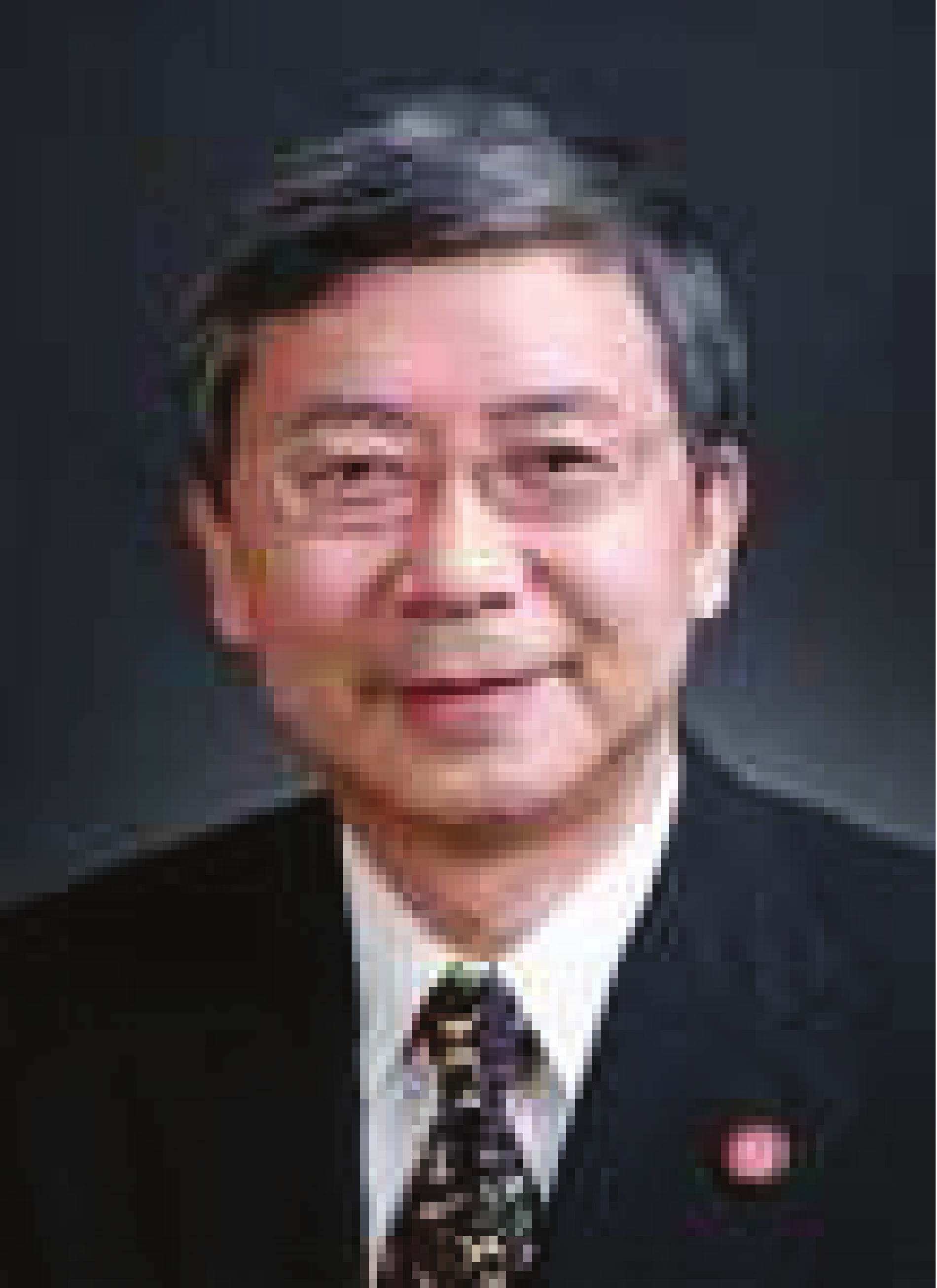}}]{Nanning Zheng}
(SM'94-F'06) graduated in 1975 from the Department of Electrical Engineering, Xi'an Jiaotong University (XJTU), received the ME degree in Information and Control Engineering from Xi'an Jiaotong University in 1981, and a Ph. D. degree in Electrical Engineering from Keio University in 1985.
He is currently a Professor and the director with the Institute of Artificial Intelligence and Robotics of Xi'an Jiaotong University. His research interests include computer vision, pattern recognition, computational intelligence, and hardware implementation of intelligent systems. Since 2000, he has been the Chinese representative on the Governing Board of the International Association for Pattern Recognition. He became a member of the Chinese Academy Engineering in 1999. He is a fellow of the IEEE.
\end{IEEEbiography}

\vspace{-60pt}

\begin{IEEEbiography}[{\includegraphics[width=1in,height=1.25in,clip,keepaspectratio]{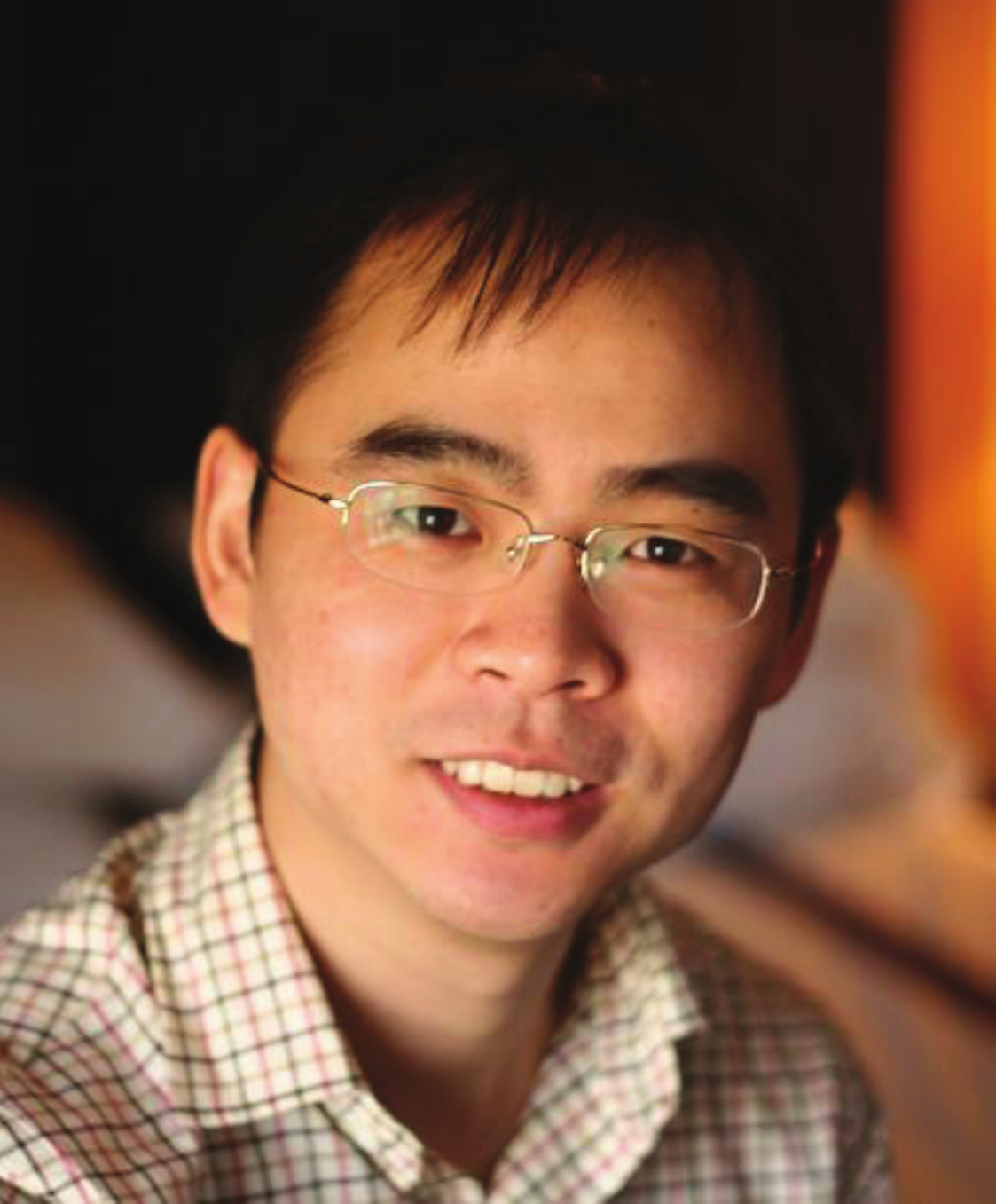}}]{Gang Hua}
(M'03-SM'11) was enrolled in the Special Class for the Gifted Young of Xi'an Jiaotong University (XJTU) in 1994 and received the B.S. degree in Automatic Control Engineering from XJTU in 1999. He received the M.S. degree in Control Science and Engineering in 2002 from XJTU, and the Ph.D. degree in Electrical Engineering and Computer Science at Northwestern University in 2006. He is currently a Principle Researcher/Research Manager at Microsoft Research. Before that, he was an Associate Professor of Computer Science at Stevens Institute of Technology. He also held an Academic Advisor position at IBM T. J. Watson Research Center between 2011 and 2014. He was a Research Staff Member at IBM Research T. J. Watson Center from 2010 to 2011, a Senior Researcher at Nokia Research Center, Hollywood from 2009 to 2010, and a Scientist at Microsoft Live Labs Research from 2006 to 2009. He is currently an Associate Editor in Chief for CVIU, and Associate Editors for IJCV, IEEE T-IP, IEEE T-CSVT, IEEE Multimedia, and MVA. He also served as the Lead Guest Editor on two special issues in TPAMI and IJCV, respectively. He is a program chair of CVPR'2019\&2022. He is an area chair of CVPR'2015\&2017, ICCV'2011\&2017, ICIP'2012\&2013\&2016, ICASSP'2012\&2013, and ACM MM 2011\&2012\&2015\&2017. He is the author of more than 150 peer reviewed publications in prestigious international journals and conferences. He holds 19 issued US patents and has 20 more US patents pending. He is the recipient of the 2015 IAPR Young Biometrics Investigator Award for his contribution on Unconstrained Face Recognition from Images and Videos, and a recipient of the 2013 Google Research Faculty Award. He is an IAPR Fellow, an ACM Distinguished Scientist, and a senior member of the IEEE.
\end{IEEEbiography}





\end{document}